\pdfoutput=1

\documentclass[10pt,twocolumn,letterpaper]{article}

\usepackage[pagenumbers]{cvpr} 

\usepackage{graphicx}
\usepackage{tablefootnote}
\usepackage{amsmath}
\usepackage{amssymb}
\usepackage{booktabs}

 \usepackage{tabularx}
\usepackage{longtable}
\usepackage{arydshln}

\usepackage{hyperref}
\usepackage{url}
\usepackage{multirow} 
\usepackage{soul}
\usepackage{changepage,threeparttable} 
\usepackage{adjustbox}
\usepackage{appendix}
\usepackage{hyperref}
\usepackage{multirow}
\usepackage{subcaption}
\usepackage{comment}
\usepackage{color}

\usepackage{enumitem}
\setlist[itemize]{leftmargin=*} 

\usepackage{array}

\newcolumntype{L}[1]{>{\raggedright\let\newline\\\arraybackslash\hspace{0pt}}m{#1}}

\usepackage{tikz}

\usepackage[accsupp]{axessibility}  

\newcounter{oldtocdepth}

\newcommand{\hidefromtoc}{%
  \setcounter{oldtocdepth}{\value{tocdepth}}%
  \addtocontents{toc}{\protect\setcounter{tocdepth}{-10}}%
}

\newcommand{\unhidefromtoc}{%
  \addtocontents{toc}{\protect\setcounter{tocdepth}{\value{oldtocdepth}}}%
}
\usepackage{bookmark}

\usepackage[capitalize]{cleveref}
\crefname{section}{Sec.}{Secs.}
\Crefname{section}{Section}{Sections}
\Crefname{table}{Table}{Tables}
\crefname{table}{Tab.}{Tabs.}

\usepackage{makecell}

\newcommand{\bz}{\mathbf{z}}
\newcommand{\bx}{\mathbf{x}}

\def\cvprPaperID{5372} 
\def\confName{CVPR}
\def\confYear{2023}

\begin{document}

\title{Re-thinking Model Inversion Attacks Against Deep Neural Networks}
\author{
Ngoc-Bao Nguyen$^*$ 
\quad
Keshigeyan Chandrasegaran$^*$ 
\quad
Milad Abdollahzadeh
\quad
Ngai-Man Cheung$^\dag$ \\
Singapore University of Technology and Design (SUTD) \\
\hspace{-0.4cm}{\tt\small thibaongoc\_nguyen@mymail.sutd.edu.sg,\{keshigeyan,milad\_abdollahzadeh,ngaiman\_cheung\}@sutd.edu.sg}
}
\maketitle
\def\thefootnote{*}
\footnotetext{Equal Contribution \hspace{3 mm} $^\dag$Corresponding Author}
\def\thefootnote{\arabic{footnote}}

\maketitle


\begin{abstract}
Model inversion (MI) attacks aim to  infer and reconstruct
private 
training data by abusing access to a model. MI attacks have raised  concerns about
the leaking of sensitive  
information (e.g. private face images used in training a face recognition system).
Recently, several algorithms for MI have been proposed to improve the attack performance.
In this work, we revisit MI,  
study two fundamental issues \textbf{pertaining to all state-of-the-art (SOTA) MI algorithms}, and propose 
solutions to these issues which lead to a significant boost in attack performance for all SOTA MI.
In particular, our contributions are two-fold: 
1) We analyze
the optimization objective of SOTA MI algorithms,
argue that the objective is sub-optimal for achieving MI, 
and propose an improved optimization objective that boosts attack performance significantly.
2) We 
analyze ``MI overfitting'', show that it would prevent reconstructed images from learning semantics of training data, and propose a novel ``model augmentation'' idea to overcome this issue. 
Our proposed solutions are simple 
and improve all SOTA MI attack accuracy significantly. E.g., in the standard CelebA benchmark, our solutions improve accuracy by \textbf{11.8\%} and achieve for the first time over 
90\% attack accuracy. 
\textbf{Our findings demonstrate that there is a clear risk of leaking sensitive information from deep learning models.} We urge serious consideration to be given to the privacy implications. 
Our code, demo, and models are available at
\textcolor{magenta}{\url{https://ngoc-nguyen-0.github.io/re-thinking_model_inversion_attacks/}}.
\end{abstract}

\hidefromtoc
\begin{figure*}[!th]
\begin{adjustbox}{width=0.99\textwidth,center}
\begin{tabular}{c}
    \includegraphics[width=0.99\textwidth]{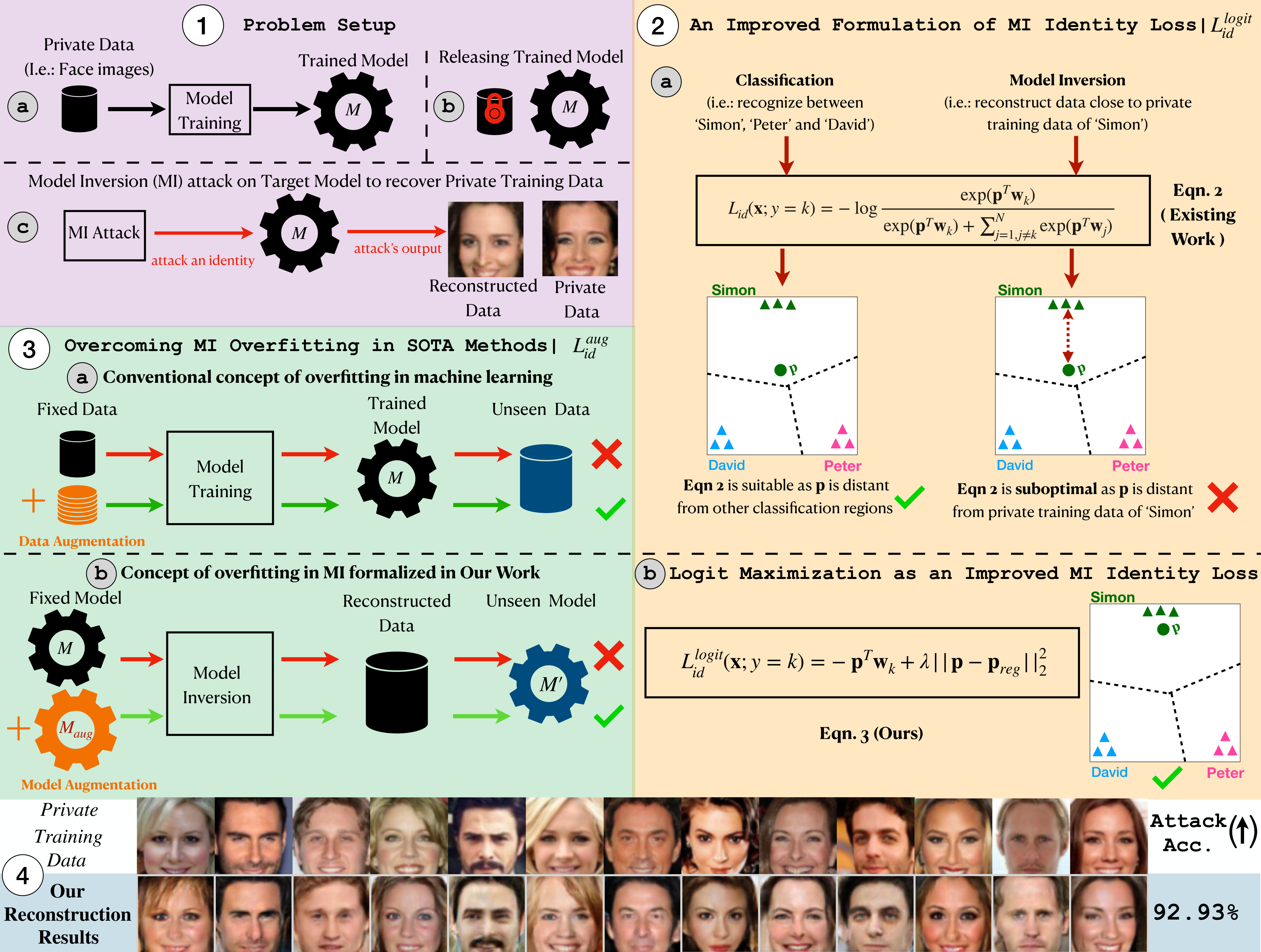}
\end{tabular}
\end{adjustbox}
\vspace{-0.35cm}
\caption{
\textit{Overview and our contributions.}
\textbf{\textcircled{\raisebox{-0.8pt}{1}}} 
We consider the problem of the Model Inversion (MI) attack to reconstruct private training data based on model parameters. 
Our work makes two foundational contributions to MI attacks.
\textbf{\textcircled{\raisebox{-0.8pt}{2}}}
First, we analyse the optimization objective of existing SOTA MI algorithms and show that they are sub-optimal. Further, we propose an improved optimization objective that boosts MI attack performance significantly (Sec \ref{sub-sec:improved_formulation_of_identity_loss}).
\textbf{\textcircled{\raisebox{-0.8pt}{3}}}
Second, we  formalize the concept of ``MI overfitting'' showing
that it prevents reconstructed images from learning identity semantics of training data. Further, we propose a novel ``model augmentation'' idea to overcome this issue (Sec \ref{sub-sec:mi_overfitting}).
\textbf{\textcircled{\raisebox{-0.8pt}{4}}}
Our proposed method significantly boosts MI attack accuracy. \Eg In the standard CelebA benchmark, our method boosts attack accuracy by \textit{11.8\%, achieving above 90\% attack accuracy for the first time in contemporary MI literature}.
}
\label{fig:overview}
\vspace{-0.4cm}
\end{figure*}


\section{Introduction}

Privacy of deep neural networks (DNNs) 
has  attracted  
considerable attention recently \cite{chabanne2017privacy,beaulieu2019privacy,sirichotedumrong2021gan,lee2022privacy,subbanna2021analysis}.
Today, DNNs are being applied in many domains involving private and sensitive datasets, e.g., healthcare, and security.
There is a growing concern 
of privacy attacks to gain knowledge of confidential datasets used in training DNNs.
One important category of privacy attacks is Model Inversion  (MI)\cite{fredrikson2014privacy,fredrikson2015model,yang2019neural,zhang2020secret,chen2021knowledge,choquette2021label,wang2021variational,zhao2021exploiting,kahla2022label}
(Fig. \ref{fig:overview}).
Given access to a model, 
MI attacks aim to infer and reconstruct features of 
the private dataset used in the training of the model.
For example, a malicious user may attack a face recognition system to reconstruct sensitive face images used in training.
Similar to previous work \cite{zhang2020secret,wang2021variational,chen2021knowledge}, we will use face recognition 
models
as the running example.

\vspace{0.05cm}
\noindent
{\bf Related Work.} 
MI attacks were first introduced in
\cite{fredrikson2014privacy}, where simple linear regression is the target of attack.
Recently, there is a fair amount of 
interest to extend MI to complex DNNs.
Most of these attacks \cite{zhang2020secret,chen2021knowledge,wang2021variational}
focus on 
 the {\em whitebox} setting and the attacker is assumed to have  complete knowledge
of the model subject to attack.
As many 
platforms
provide downloading of entire 
trained DNNs 
for users \cite{zhang2020secret,chen2021knowledge},
whitebox attacks are important. 
\cite{zhang2020secret} proposes Generative
Model Inversion (GMI) attack, where 
generic 
public information is leveraged to learn
a distributional prior via generative adversarial networks
(GANs)
\cite{goodfellow2020generative, tran2021data}, and this prior is used to guide reconstruction of private training samples.
\cite{chen2021knowledge} proposes 
Knowledge-Enriched Distributional Model Inversion (KEDMI), where an  inversion-specific GAN is trained by leveraging  
knowledge provided by the target model.
\cite{wang2021variational} proposes 
Variational Model Inversion (VMI), where  a probabilistic interpretation of MI leads to a variational objective for the attack.
KEDMI and VMI achieve SOTA attack performance
(See Supplementary \ref{supp-sec:related_work} for further discussion of related work).

{\bf In this paper}, we revisit SOTA MI, study two issues pertaining to all SOTA MI and propose solutions to these issues that are complementary and applicable to all SOTA MI
(Fig. \ref{fig:overview}).
In particular, despite the range of approaches proposed in recent works, common and central to all these approaches 
is an {\em inversion step} which formulates
 reconstruction of training samples  as an optimization.
The optimization objective in the inversion step  involves the  {\em identity loss}, which is the {\em same} for all SOTA MI and is formulated as the negative log-likelihood for the reconstructed samples under the model being attacked.
While  ideas have been proposed to advance  other aspects of MI, {\em effective design of the identity loss has not been studied}.

\noindent
To address this research gap, our work studies subtleties of identity loss in all SOTA MI, analyzes the issues and proposes improvements that boost the performance of all SOTA significantly.
In summary, our contributions are as follows:

\begin{itemize}
    \item  We analyze existing identity loss, argue that it could be sub-optimal for MI, and propose an improved identity loss that aligns better with the goal of MI (Fig. \ref{fig:overview} \textcircled{\raisebox{-0.8pt}{2}}).
    \item We formalize the concept of {\em MI overfitting}, analyze its impact on MI and propose a novel solution based on {\em model augmentation}. Our idea is inspired by the  conventional issue of overfitting in model training and data augmentation as a solution to alleviate the issue (Fig. \ref{fig:overview} \textcircled{\raisebox{-0.8pt}{3}}).
    \item We conduct extensive experiments to demonstrate that our solutions can improve SOTA MI algorithms (GMI \cite{zhang2020secret}, KEDMI \cite{chen2021knowledge}, VMI \cite{wang2021variational}) significantly.
    Our solutions achieve for the first time over 90\% attack accuracy under standard CelebA benchmark (Fig. \ref{fig:overview} \textcircled{\raisebox{-0.8pt}{4}}).
\end{itemize}

Our work sounds  alarm over the rising threats of MI attacks, and urges more attention on measures against the leaking of private information from DNNs.

\section{General Framework of SOTA MI Attacks}
\label{sub-sec:unified_view_of_sota_mi}

{\bf Problem Setup.} 
In MI, an attacker abuses access to a model
$M$ trained on a private dataset $\mathcal{D}_{priv}$.
The attacker can access  $M$, but $\mathcal{D}_{priv}$ is not intended to be shared.
The goal of MI is to infer information about private samples in $\mathcal{D}_{priv}$.
In existing work, for the desired class (label) $y$,
MI is formulated as the reconstruction of 
an input $\bx$ which is most likely classified into  $y$ by the model $M$.
For instance, if the problem involves inverting a facial recognition model, given the desired identity, MI is formulated as the reconstruction of facial images that are most likely to be recognized as the desired identity. The model subject to MI attacks is called {\em target model}.
Following previous works \cite{zhang2020secret,chen2021knowledge,wang2021variational},
we focus on {\em whitebox} MI attack, where the attacker is assumed to have complete access to the target model.
For high-dimensional data such as facial images, this reconstruction problem is ill-posed.
Consequently, various SOTA MI methods have been proposed recently to constrain the search space to the manifold of meaningful and relevant images
using a GAN: using a GAN trained on some public dataset $\mathcal{D}_{pub}$ ~\cite{zhang2020secret}, using an inversion-specific GAN~\cite{chen2021knowledge}, and defining variational inference in latent space of GAN~\cite{wang2021variational}.

\begin{table}
    \centering
    \caption{Categorizing SOTA MI attacks based on their difference in latent code distribution and prior loss. $p_\mathtt{GAN}(\bz)$ is a GAN prior. $G$ and $D$ are generator and discriminator of a GAN.
    }
    \resizebox{\columnwidth}{!}{%
    \begin{tabular}{lcc}
    \toprule
    \thead{Method} & \thead{Latent distribution ${q(\bz)}$} & \thead{Prior loss ${L}_{prior}$} \\
    \midrule
    
    {GMI}~\cite{zhang2020secret} & \makecell{Point estimate $\delta(\bz-\bz_0)$} & \makecell{ $-D(G(\bz))$} \\
    \midrule
    
    {KEDMI}~\cite{chen2021knowledge} & \makecell{Gaussian $ \boldsymbol{\mathcal{N}}(\boldsymbol{\mu},\boldsymbol{\Sigma})$} & \makecell{$-\log{D(G(\bz))}$} \\
    \midrule
    
    {VMI}~\cite{wang2021variational} & \makecell{Gaussian $\boldsymbol{\mathcal{N}} (\boldsymbol{\mu},\boldsymbol{\Sigma})$ {\em or} \\ Normalizing Flow~\cite{kingma2018glow}} & \makecell{Distance \textit{w.r.t.} GAN prior \\ $D_\mathtt{KL}(q(\bz)||p_\mathtt{GAN}(\bz))$} \\
    \bottomrule
    
    \end{tabular}}
    \label{tab:SOTA_MI_Comparison}
\end{table}

Despite the differences in various SOTA MI, common and central to all these methods is an {\em inversion step}  
--called {\em secret revelation} in~\cite{zhang2020secret}--, which performs the following optimization:
\begin{equation}
\label{eqn:overall_mi_objective}
    q^*(\bz) = \arg \min_{q(\bz)} 
    \mathbb{E}_{\bz \sim q(\bz)} \{{L}_{id}(\bz;y,M) + \lambda {L}_{prior}(\bz)\}
\end{equation}
Here
${L}_{id}(\bz;y,M)=-\log \mathbb{P}_M(y|G(\bz))$
is referred to as {\em identity loss} in MI \cite{zhang2020secret}, which guides the reconstruction of $\bx=G(\bz)$ that is most likely to be recognized by model $M$ as identity $y$, and
${L}_{prior}$ is some prior loss, and 
$q^*(\bz)$ is the optimal distribution of latent code used to generate inverted samples by GAN ($\bx=G(\bz)$; $\bz \sim q^*(\bz)$). 
Importantly, all  SOTA MI methods use the {\em same} identity loss ${L}_{id}(\bz;y,M)$, although they have different assumption 
about $q(\bz)$ and the prior loss ${L}_{prior}$ (see Table \ref{tab:SOTA_MI_Comparison} and  Supplementary for more details on each algorithm).
While advances observed by improving $q(\bz)$ and ${L}_{prior}$, \emph{\boldmath\bfseries the design of more effective ${L}_{id}$ has been left unnoticed} in all SOTA MI algorithms.
Therefore, our work instead focuses on ${L}_{id}$, analyzes issues, and proposes improvement for ${L}_{id}$ that can lead to a performance boost in all SOTA MI.
To simplify notations, we denote ${L}_{id}(\bz;y,M)$  by  ${L}_{id}(\bx;y)$ when appropriate, where  $\bx=G(\bz)$ is the reconstructed image.

\section{A Closer Look at Model Inversion Attacks}

\subsection{An Improved Formulation of MI Identity Loss}
\label{sub-sec:improved_formulation_of_identity_loss}

\begin{figure*}[!th]
\begin{adjustbox}{width=0.99\textwidth,center}
\begin{tabular}{c}
    \includegraphics[width=0.99\textwidth]{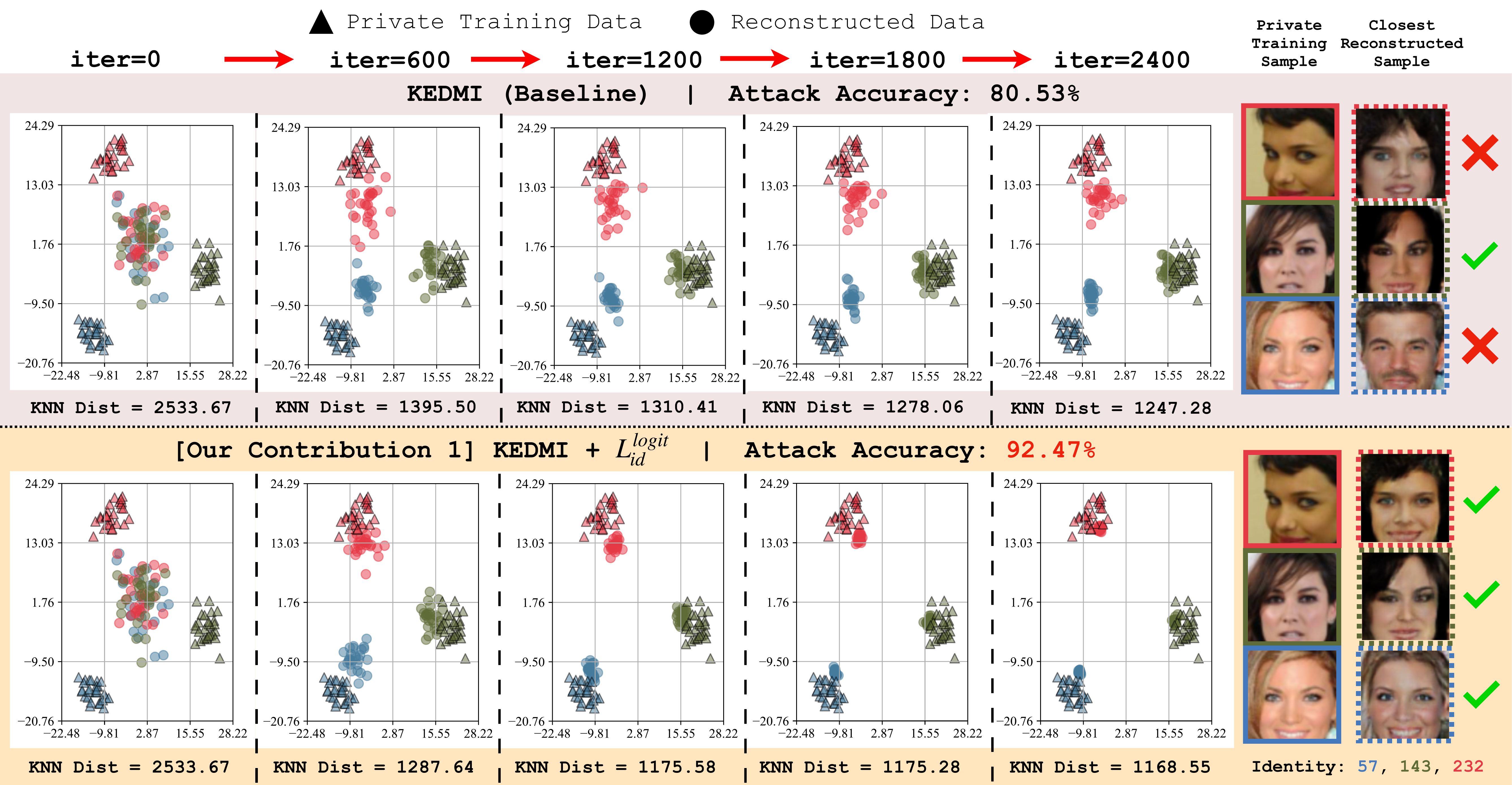}
\end{tabular}
\end{adjustbox}
\vspace{-0.35cm}
\caption{
Visualization of the penultimate layer representations ($\mathcal{D}_{priv}$ = CelebA \cite{liu2015deep}, $\mathcal{D}_{pub}$ = CelebA \cite{liu2015deep}, Target Model = IR152 \cite{chen2021knowledge}, Evaluation Model = face.evoLve \cite{yu_know_you_at_one_glance}, Inversion iterations = 2400) for private training data and reconstructed data using KEDMI \cite{chen2021knowledge}.
Following the exact evaluation protocol in \cite{chen2021knowledge}, we use face.evoLve \cite{yu_know_you_at_one_glance} to extract representations.
We show results for 3 randomly chosen identities.
We include KNN distance (for different iterations) and final attack accuracy following the protocol in \cite{chen2021knowledge}.
For each identity, we also include randomly selected private training data and the closest reconstructed sample at iteration=2400.
\textbf{\textcircled{\raisebox{-0.8pt}{1}}
Identity loss in SOTA MI methods \cite{zhang2020secret, chen2021knowledge, wang2021variational}
(Eqn. \ref{eqn:id_loss_existing_works_decomposed})
is sub-optimal for MI (Top).}
Using penultimate representations during inversion, we observe 2 instances (\eg target identity {\bf \color[HTML]{4579BD} 57} and {\bf \color[HTML]{606c38} 143}) 
where KEDMI \cite{chen2021knowledge} (using Eqn. \ref{eqn:id_loss_existing_works_decomposed} for identity loss) is unable to reconstruct data close to private training data.
Hence, private and reconstructed facial images are qualitatively different. 
\textbf{\textcircled{\raisebox{-0.8pt}{2}} 
Our proposed identity loss, ${L}_{id}^{logit}$ (Eqn. \ref{eqn:our_id_loss}), 
can 
effectively guide the reconstruction of data close to 
private training data 
(Bottom)}.
This can be clearly observed using both penultimate layer representations and KNN distances for all 3 target classes {\bf \color[HTML]{4579BD} 57}, {\bf \color[HTML]{606c38} 143} and {\bf \color[HTML]{E63946} 252}.
We show similar results using additional MI algorithms (GMI \cite{zhang2020secret}, VMI \cite{wang2021variational}) and target classifiers (face.evoLve, VGG16) in Supplementary Figures 
\ref{fig-supp:gmi_penultimate_visualization},
\ref{fig-supp:vmi_penultimate_visualization} and 
\ref{fig:kedmi_penultimate_visualization_face_evolve_target}.
Best viewed in color.
}
\label{fig:ked_mi_penultimate_visualization}
\vspace{-0.5cm}
\end{figure*}



In this section, we discuss our first contribution and 
take a closer look at
the optimization objective of \textit{identity loss}, 
${L}_{id}(\bx;y)$.
Existing SOTA MI methods, namely GMI \cite{zhang2020secret}, KEDMI \cite{chen2021knowledge} and VMI \cite{wang2021variational} formulate the identity loss as an optimization to minimize the negative log likelihood of an identity under model parameters (\ie cross-entropy loss). 
Particularly, the 
${L}_{id}(\bx;y)$
introduced in Eqn. \ref{eqn:overall_mi_objective} for an inversion targeting class $k$ can be re-written as follows:

\def\pvector{\textbf{p}}
\def\wvector{\textbf{w}}

\vspace{-0.6cm}
\begin{equation} 
\label{eqn:id_loss_existing_works_decomposed}
{L}_{id}(\bx;y=k) =  - \log \frac{ \exp({\pvector^T\wvector_k}) } { \exp({\pvector^T\wvector_k}) + \sum_{j=1, j \neq k}^{N} \exp({\pvector^T\wvector_j}) }
\end{equation}
where $\pvector$ refers to penultimate layer activations \cite{muller, pmlr-v162-chandrasegaran22a} for sample $\bx$ and $\wvector_i$ refers to the last layer weights for the $i^{th}$ class 
\footnote{
$\pvector$ is concatenated with 1 at the end to include bias as $\wvector_i$ includes biases at the end.}{in target model $M$}. 

\vspace{0.2cm}
\noindent
\textbf{Existing identity loss (Eqn. \ref{eqn:id_loss_existing_works_decomposed}) used in SOTA MI methods \cite{zhang2020secret, chen2021knowledge, wang2021variational} is sub-optimal for MI}
(Fig. \ref{fig:overview}
\textcircled{\raisebox{-0.8pt}{2}}).
Although the optimization in Eqn. \ref{eqn:id_loss_existing_works_decomposed} accurately captures the essence of a classification problem (\eg face recognition), 
we postulate that such formulation is sub-optimal for MI.
We provide our intuition through the lens of penultimate layer activations, $\pvector$
(Fig. \ref{fig:overview}
\textcircled{\raisebox{-0.8pt}{2}}).
In a classification setting, 
the main expectation for $\pvector$ is to be sufficiently discriminative for class $k$ 
(\eg recognize between `Peter', `Simon' and `David'). 
This objective can be achieved by both maximizing $\exp({\pvector^T \wvector_k})$ and/or minimizing $\sum_{j=1, j \neq k}^{N} \exp({\pvector^T \wvector_j})$ in Eqn. \ref{eqn:id_loss_existing_works_decomposed}.
On the contrary, \textit{the goal of MI is to reconstruct training data}.
That is, in addition to $\pvector$ being sufficiently discriminative for class $k$, successful inversion also requires $\pvector$ to be close to the 
training data representations for class $k$ represented by $\wvector_k$ 
(\ie an inversion targeting `Simon' needs to reconstruct a sample 
close to the private training data of `Simon'; Fig. \ref{fig:overview}
\textcircled{\raisebox{-0.8pt}{2}}).
Specifically, we argue that MI requires a lot more attention on maximizing $\exp({\pvector^T\wvector_k})$ compared to minimizing $\sum_{j=1, j \neq k}^{N} \exp({\pvector^T\wvector_j})$ in Eqn. \ref{eqn:id_loss_existing_works_decomposed}.

Motivated by this hypothesis, we conduct an analysis to investigate the proximity between private training data and reconstructed data in SOTA MI methods using penultimate layer representations \cite{pmlr-v162-chandrasegaran22a, muller, shen2021, lukasik20a}.
Particularly, our analysis using KEDMI \cite{chen2021knowledge} (SOTA) shows several instances where using Eqn. \ref{eqn:id_loss_existing_works_decomposed} for identity loss is unable to reconstruct data close to the private training data.
We show this in Fig. \ref{fig:ked_mi_penultimate_visualization} (top row). Consequently, our analysis motivates the search for an improved identity loss focusing on maximizing $\exp({\pvector^T\wvector_k})$ for MI.

\vspace{0.2cm}
\noindent
\textbf{Logit Maximization as an improved MI identity loss.}
In light of our analysis / observations above, 
we propose to directly maximize the logit, ${\pvector^T\wvector_k}$, instead of maximizing the log likelihood of class $k$ for MI. 
Our proposed identity loss objective is shown below:

\vspace{-0.5cm}
\begin{equation} 
\label{eqn:our_id_loss}
{L}_{id}^{logit}(\bx;y=k) =  - { {\pvector^T\wvector_k} } + \lambda ||\pvector - \pvector_{reg}||^2_2
\end{equation}

\noindent
where $\lambda(>0)$ is a hyper-parameter and $\pvector_{reg}$ is used for regularizing $\pvector$.
Particularly, 
if the regularization in Eqn. \ref{eqn:our_id_loss} is omitted and hence 
$||\pvector||$ is unbounded, a crude simplified way to solve Eqn. \ref{eqn:our_id_loss} is to maximize $||\pvector||$. Hence, 
we use $\pvector_{reg}$ to regularize $\pvector$.
Given that the attacker has no access to private training data, 
we estimate $\pvector_{reg}$ 
by a simple method using {\em public} data (See Supplementary \ref{sub:pvector}).
We remark that $\pvector=M^{\mathtt{pen}}(\bx)$ where $\bx=G(\bz)$  
and $M^{\mathtt{pen}}()$ operator returns the penultimate layer representations for a given input.

Our analysis shows that our proposed identity loss, ${L}_{id}^{logit}$ (Eqn. \ref{eqn:our_id_loss}), can significantly improve reconstruction of private training data compared to existing identity loss used in SOTA MI algorithms \cite{zhang2020secret, chen2021knowledge, wang2021variational}. This can be clearly observed using both penultimate layer representations and KNN distances in Fig. 
 \ref{fig:ked_mi_penultimate_visualization} (bottom row).
Here KNN Dist refers to the shortest Euclidean feature distance from a reconstructed image to private training images for a given identity \cite{zhang2020secret, chen2021knowledge}.
Our proposed ${L}_{id}^{logit}$ can be easily plugged in to all existing SOTA MI algorithms 
by replacing ${L}_{id}$ with our proposed ${L}_{id}^{logit}$ in Eqn. \ref{eqn:overall_mi_objective} (in the inversion step) with minimal computational overhead.

\subsection{Overcoming MI Overfitting in SOTA methods}
\label{sub-sec:mi_overfitting}

In this section, we discuss our second contribution.
In particular, we formalize a concept of \textit{MI overfitting}, observe its considerable impacts even in SOTA MI methods \cite{zhang2020secret, chen2021knowledge, wang2021variational}, and propose a new, simple solution to overcome this issue
(Fig. \ref{fig:overview}
\textcircled{\raisebox{-0.8pt}{3}}).
To better discuss our MI overfitting concept, 
we first review the conventional concept of overfitting in machine learning: 
Given the fixed training dataset and our goal of learning a model, conventionally, overfitting is defined as instances which during model learning (training stage),
the model fits too
closely to the training data and adapts to the random variation
and noise of training data, failing to adequately
learn the semantics of the training data \cite{santos2022avoiding, yu2022understanding, abdollahzadeh2021revisit, zhao2022fewshot, teo2022fair}. 
As the model lacks semantics of training data, it could be observed that the model performs poorly under unseen data
(Fig. \ref{fig:overview}
\textcircled{\raisebox{-0.8pt}{3}}
\textcircled{\raisebox{-0.8pt}{a}}).

\begin{figure}[!h]
\label{fig:mi_overfitting}
\begin{adjustbox}{width=0.96\columnwidth,left}
\begin{tabular}{c}
    \includegraphics[width=0.99\columnwidth]{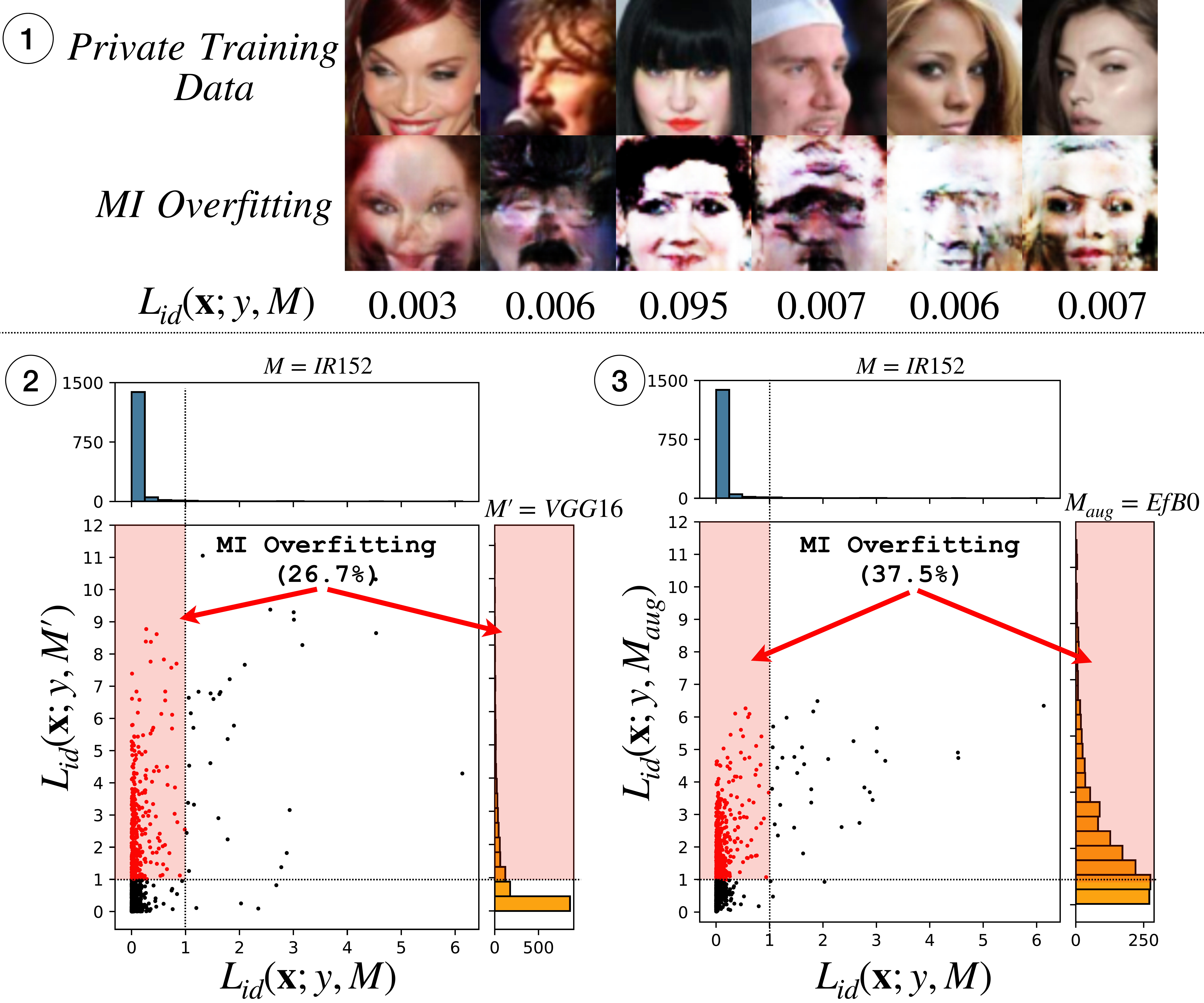}
\end{tabular}
\end{adjustbox}
\vspace{-0.35cm}
\caption{
Qualitative / Quantitative studies to demonstrate MI overfitting in SOTA methods.
We demonstrate this observation using KEDMI \cite{chen2021knowledge}. 
We use $\mathcal{D}_{priv}$ = CelebA \cite{liu2015deep}, $\mathcal{D}_{pub}$ = CelebA \cite{liu2015deep} and $M$ = IR152 \cite{chen2021knowledge}.
\textbf{\textcircled{\raisebox{-0.8pt}{1}} We show qualitative results to illustrate MI overfitting.} 
We show 6 identities, top: private data, bottom: reconstructed data from $M$.
The reconstructed samples have fit too closely to $M$ during inversion resulting in  samples with lack of identity semantics. 
Particularly, we remark that these samples have very low identity loss under the target model $M$.
\textbf{\textcircled{\raisebox{-0.8pt}{2}} Quantitative results validating the prevalence of MI overfitting in SOTA MI methods.
}
We use an additional target classifier $M'$ = VGG16 released by \cite{chen2021knowledge, zhang2020secret} to quantitatively verify the presence of MI overfitting using identity loss.
For 1,500 reconstructed samples from $M$, we visualize their identity loss \wrt $M$ and $M'$ in the 
scatter plot and respective histograms.
Particularly, we find that there are 26.7\% of samples with low identity loss under the target model $M$, but large identity loss under unseen VGG16 model $M'$, 
hinting that these samples might lack identity semantics. This result shows that 
MI overfitting is a considerable issue even in SOTA MI methods.
Note that VGG16 is used here only for analysis and is not part of our solution, as private data is not available.
\textbf{\textcircled{\raisebox{-0.8pt}{3}}
Model Augmentation to alleviate MI overfitting during inversion.} We repeat the above analysis, with $M'$ = VGG16 replaced by $M_{aug}$ = EfficientNet-B0. Importantly, $M_{aug}$ is trained by {\em public data using knowledge distillation} \cite{hinton2015distilling}.
We similarly observe samples with large identity loss under $M_{aug}$.
}
\label{fig:overfit_visualization}
\vspace{-0.2cm}
\end{figure}

\vspace{0.02cm}
\noindent
\textbf{Overfitting in MI}.
We formalize the concept of overfitting in MI
(Fig. \ref{fig:overview}
\textcircled{\raisebox{-0.8pt}{3}}
\textcircled{\raisebox{-0.8pt}{b}}).
Given the fixed (target) model and our goal of learning reconstructed samples, we define
MI overfitting as instances which during model inversion,
the reconstructed samples fit too
closely to the target model and adapt to the random variation
and noise of the target model parameters, failing to adequately
learn semantics of the identity. 
As these reconstructed samples lack identity semantics, it could be observed that they perform poorly under another unseen
model.

\vspace{0.15cm}
\noindent
{\bf Analysis.}
In what follows, we discuss our  analysis to demonstrate MI overfitting  and understand its impact in SOTA.
See Fig.~\ref{fig:overfit_visualization} for analysis setups and results. In particular, 
 in Fig.~\ref{fig:overfit_visualization} \textcircled{\raisebox{-0.8pt}{1}},
we show  some
 reconstructed samples 
which achieve low identity loss under the target model $M$, yet they lack identity semantics.
In Fig.~\ref{fig:overfit_visualization} \textcircled{\raisebox{-0.8pt}{2}},
we show that  for a considerable percentage of 
reconstructed samples from target model $M$ with  
low identity loss under $M$, 
their identity loss under another unseen model $M'$ is large as shown in the scatter plot and histograms, hinting that these samples might have suffered from MI overfitting and lack identity semantics.
We note that  the identity loss under $M'$ is obtained by feeding the reconstructed sample into $M'$ in a forward pass.
We also note that SOTA KEDMI  \cite{chen2021knowledge} is used in this analysis but the issue persists in \cite{zhang2020secret, wang2021variational}.

\vspace{0.15cm}
\noindent
{\bf Our proposed solution to MI overfitting.}
We
propose a novel solution based on {\em model augmentation}.
Our idea is inspired by the conventional issue of
overfitting in model training and data augmentation as a solution
to alleviate the issue.
In particular, for conventional overfitting, augmenting the training dataset could alleviate the issue
\cite{alexnet}.
Therefore, we hypothesize that by augmenting the target model we can alleviate MI overfitting.

Specifically, 
we propose to apply knowledge distillation (KD) \cite{hinton2015distilling}, with target model $M_t$ as the teacher, to train {augmented models} $M_{aug}^{(i)}$. Importantly, as we do not have access to the private data, during KD, each $M_{aug}^{(i)}$ is trained on the {\em public dataset} to match its output to the output of $M_t$. 
We select different network architectures for $M_{aug}^{(i)}$ and they are different from $M_t$ (Detailed discussion in the Supplementary \ref{sub:no_Maug} and \ref{sub:architecture_Maug}).
After performing KD, we apply  $M_{aug}^{(i)}$ together with the target model $M_t$ in the inversion step and compute the
identity loss (with model augmentation): 
\begin{equation}
\begin{aligned}
{L}_{id}^{aug}(\bx;y) &= 
   \gamma_t \cdot {L}_{id}(\bx;y,M_t) \\ 
   &+ \gamma_{aug} \cdot \sum_{i=1}^{N_{aug}} {L}_{id}(\bx;y,M_{aug}^{(i)}) 
\end{aligned}
\label{eq:augmentation}
\end{equation}

Here, $\gamma_t$ and $\gamma_{aug}$ are two hyper-parameters. In particular, we use $\gamma_t = \gamma_{aug} = \frac{1}{N_{aug}+1}$, where $N_{aug}$ is the number of augmented models.
${L}_{id}^{aug}$
in 
Eqn. \ref{eq:augmentation} is used to replace ${L}_{id}$
in the inversion step in Eqn. 
\ref{eqn:overall_mi_objective}.
Furthermore, our proposed 
${L}_{id}^{logit}$
in Eqn. \ref{eqn:our_id_loss} can be used in  Eqn. \ref{eq:augmentation} to combine the improvements. See details in Supplementary \ref{sub:combination}.

In Fig.~\ref{fig:overfit_visualization} \textcircled{\raisebox{-0.8pt}{3}}, we analyze the performance of $M_{aug}^{(i)}$. 
Similar to using the unseen model 
$M'$, we observe samples with large identity loss under $M_{aug}^{(i)}$, suggesting that samples with MI overfitting perform poorly 
under $M_{aug}^{(i)}$ as these samples lack identity semantic.

\begin{table*}[!h]
\caption{
We follow the exact the experiment setups in \cite{chen2021knowledge} 
for GMI\cite{zhang2020secret} and KEDMI\cite{chen2021knowledge}.
For VMI \cite{wang2021variational}, we follow the exact experiment setups in \cite{wang2021variational}.
In total, we conduct 72 experiments spanning 18 setups 
to demonstrate the effectiveness of our proposed method.}

\centering
\resizebox{0.87\linewidth}{!}{%
\begin{tabular}{cccccc }
\hline
\textbf{Method} & \textbf{Private Dataset} & \textbf{Public Dataset} & \textbf{Target model} & \textbf{Evaluation Model} & \textbf{Model Augmentation} \\ \hline
\multirow{5}{*}{\begin{tabular}[c]{@{}l@{}}GMI \cite{zhang2020secret} / \\ KEDMI \cite{chen2021knowledge}\end{tabular}} & \multicolumn{1}{l }{\multirow{3}{*}{CelebA \cite{liu2015deep}}} & \multirow{2}{*}{CelebA /} & \multirow{3}{*}{\begin{tabular}[c]{@{}l@{}}VGG16 \cite{simonyan2014very} / \\ IR152 \cite{he2016deep} / \\ face.evoLve \cite{cheng2017know}\end{tabular}} & \multirow{3}{*}{face.evoLve} & \multirow{4}{*}{\begin{tabular}[c]{@{}c@{}} EfficientNet-B0 \cite{tan2019efficientnet},\\ EfficientNet-B1 \cite{tan2019efficientnet}, \\ EfficientNet-B2 \cite{tan2019efficientnet} \end{tabular}} \\
 & \multicolumn{1}{c}{} & \multirow{2}{*}{FFHQ \cite{karras2019style} } & & & \\
 & \multicolumn{1}{c}{} & & & & \\ \cline{2-5}
 & \multicolumn{1}{c}{CIFAR-10 \cite{krizhevsky2010cifar}} & CIFAR-10 & VGG16 & ResNet-18 \cite{he2016deep} & \\ \cline{2-6} 
 & \multicolumn{1}{c}{MNIST \cite{lecun1998gradient}} & MNIST & CNN(Conv3) & CNN(Conv5) & CNN(Conv2), CNN(Conv4) \\ \hline
\multirow{2}{*}{VMI \cite{wang2021variational}} & \multicolumn{1}{c}{CelebA} & CelebA & ResNet-34 \cite{he2016deep} & IR-SE50 \cite{deng2019arcface} & \begin{tabular}[c]{@{}l@{}} EfficientNet-B0,\\ EfficientNet-B1, \\ EfficientNet-B2 \end{tabular} \\ \cline{2-6} 
 & MNIST & EMNIST\cite{cohen2017emnist} & ResNet-10 & ResNet-10 & CNN(Conv2), CNN(Conv4) \\ \hline
\end{tabular}
}
\label{tab:setups}
\end{table*}

\section{Experiments}

In this section, we evaluate the performance of the proposed method in recovering a representative input from the target model, against current SOTA methods: GMI~\cite{zhang2020secret}, VMI~\cite{wang2021variational}, and KEDMI~\cite{chen2021knowledge}.
More specifically, as our proposed method identifies two major limitations in current $L_{id}(\bx;y)$ --used commonly in all SOTA MI approaches-- we will evaluate the improvement brought by our improved identity loss $L_{id}^{logit}$, and model augmentation $L_{id}^{aug}$ for all SOTA MI approaches.

\subsection{Experimental Setup}
In order to have a fair comparison, when evaluating our method against each SOTA MI approach, we follow the exactly same experimental setup of that approach. In what follows, we discuss the details of these setups.

\vspace{0.15cm}
\noindent
{\bf Dataset.}
Following previous works, we evaluate the proposed method on different tasks: face recognition and digit classification is used for comparison with all three SOTA approaches, 
and image classification is used for comparison with GMI~\cite{zhang2020secret}, and KEDMI~\cite{chen2021knowledge}.
For the face recognition task, we use CelebA dataset~\cite{liu2015deep} that includes celebrity images, and the FFHQ dataset~\cite{karras2019style} which contains images with larger variation in terms of background, ethnicity, and age.
The MNIST handwritten digits dataset~\cite{lecun1998gradient} is used for 
digit classification. 
We utilize the CIFAR-10 dataset~\cite{krizhevsky2010cifar} for image classification.

\vspace{0.15cm}
\noindent
{\bf Data Preparation Protocol.} Following previous SOTA approaches~\cite{zhang2020secret, chen2021knowledge, wang2021variational}, we split each dataset into two disjoint parts: one part is used as private dataset $\mathcal{D}_{priv}$ for training target model, and another part is used as a public dataset $\mathcal{D}_{pub}$ to extract the prior information. Most importantly, {\em throughout all experiments, public dataset $\mathcal{D}_{pub}$ has no class intersection with private dataset $\mathcal{D}_{priv}$ used for training target model.} Note that this is essential to make sure that adversary uses $\mathcal{D}_{pub}$ only to gain prior knowledge about features that are general to that task (i.e., face recognition), and does not have access to information about class-specific and private information used for training target model.

\vspace{0.15cm}
\noindent
{\bf Models.}
Following previous works, we implement several different models with varied complexities.
As GMI~\cite{zhang2020secret} and KEDMI~\cite{chen2021knowledge} use exactly similar model architecture in experiments, for comparison with these two algorithms, we use the same models. More specifically, for face recognition on CelebA and FFHQ, we use VGG16~\cite{simonyan2014very}, IR152~\cite{he2016deep}, and face.evoLve~\cite{cheng2017know}--as SOTA face recognition model. For digit classification on MNIST, we use a CNN with 3 convolutional layers and 2 pooling layers. Finally, for image classification, following~\cite{chen2021knowledge} we use VGG16~\cite{simonyan2014very}. For a fair comparison with VMI, we follow its design in~\cite{wang2021variational} and use ResNet-34 for face recognition CelebA, and ResNet-10 for digit classification on MNIST. 
The details of the target models, augmented models and datasets used in experiments are summarized in Table \ref{tab:setups}. 
We remark that when comparing our proposed method with each of the SOTA MI approaches, we use exactly the same target model and GAN for both SOTA and our approach.

\vspace{0.15cm}
\noindent
{\bf Evaluation Metrics.}
To evaluate the performance of a MI attack, we need to assess whether the reconstructed image
exposes private information about a target label/identity. In this work, following the literature, we conduct both qualitative evaluations by visual inspection, and quantitative evaluations using different metrics, including:
\begin{itemize}
 \item {\bf Attack Accuracy (Attack Acc).} Following ~\cite{zhang2020secret, chen2021knowledge, wang2021variational}, we use an {\em evaluation model} that predicts the label/identity of the reconstructed image. 
 Similar to previous works, the evaluation model is different from the target model (different structure/ initialization seed), but it is trained on the same private dataset (see Table \ref{tab:setups}). Intuitively, considering a highly accurate evaluation model, it can be viewed as a proxy for human inspection ~\cite{zhang2020secret}. Therefore, if the evaluation model infers high accuracy on reconstructed images, it means these images are exposing private information about the private dataset, i.e. high attack accuracy.

 \item{\bf K-Nearest Neighbors Distance (KNN Dist).} KNN Dist indicates the distance between the reconstructed image for a specific label/id and corresponding images in the private training dataset. More specifically, it measures the shortest feature distance from the reconstructed image to the real images in the private dataset, given a class/id. It is measured as $l_2$ distance between two images in the feature space, i.e., the penultimate layer of the evaluation model.
 
\end{itemize}

\begin{figure}[h]
\label{fig:samples}
\vspace{-0.55cm}
\begin{adjustbox}{width=0.99\columnwidth,left}
\begin{tabular}{c}
 \includegraphics[width=0.99\columnwidth]
 {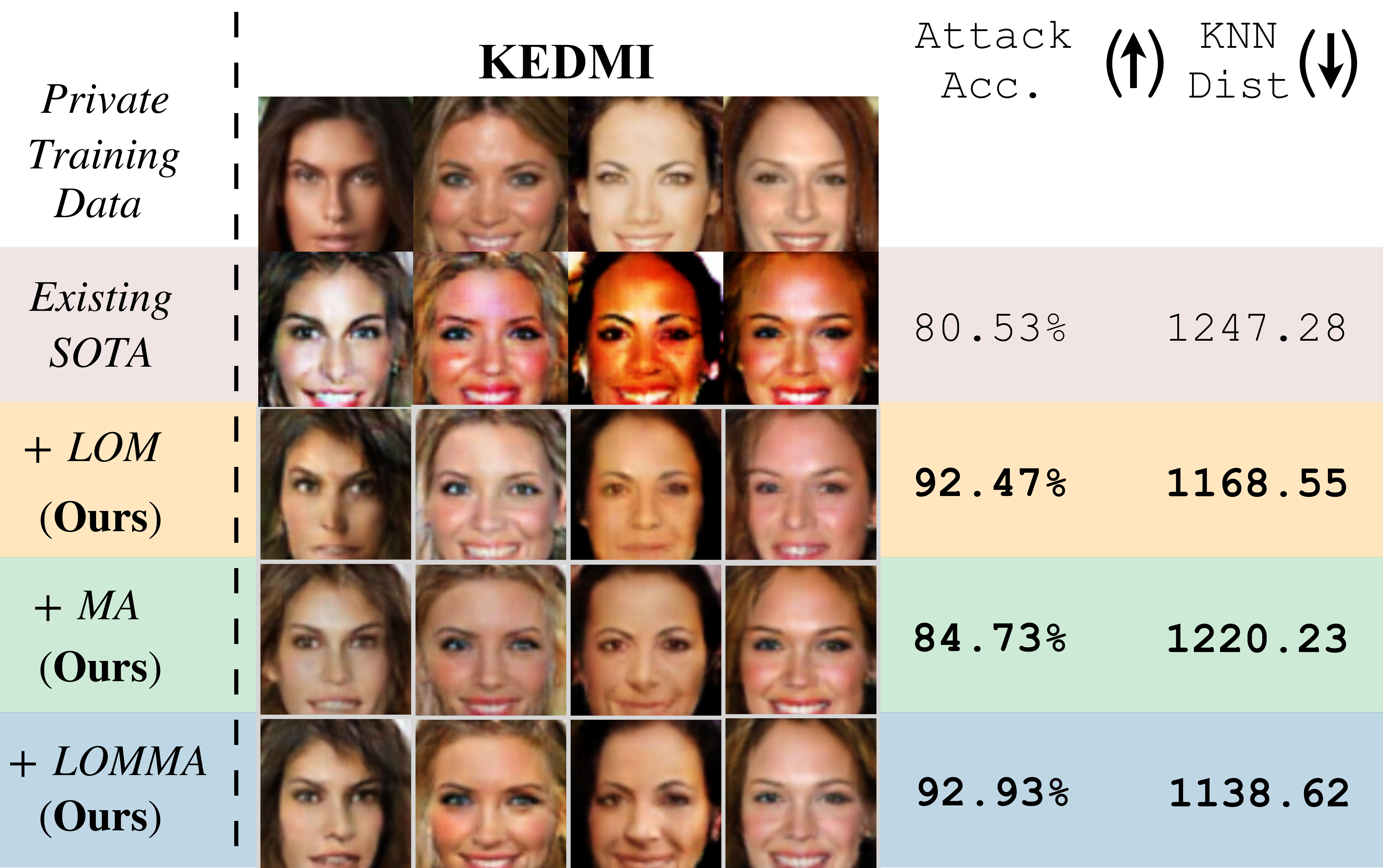}
\end{tabular}
\end{adjustbox}
\vspace{-0.35cm}
\caption{
Qualitative / Quantitative (Top1 Attack Acc., KNN Dist) results to demonstrate the efficacy of our proposed method.
We use KEDMI \cite{chen2021knowledge} (SOTA), $\mathcal{D}_{priv}$ = CelebA \cite{liu2015deep}, $\mathcal{D}_{pub}$ = CelebA \cite{liu2015deep} and $M$ = IR152 \cite{he2016deep}.
As one can observe, our proposed method achieves better reconstruction of private data both visually and quantitatively (validated by KNN results) resulting in a significant boost in attack performance.
}
\label{fig:kedmi_samples}
\vspace{-0.3cm}
\end{figure}

\begin{table}
\caption{
We report the results for KEDMI and GMI for IR152, face.evoLve and VGG16 target model. Following exact experiment setups in \cite{chen2021knowledge}, here $\mathcal{D}_{priv}$ = CelebA, $\mathcal{D}_{pub}$ = CelebA, evaluation model = face.evoLve.
We report top 1 accuracies, the improvement compared to the SOTA MI (Imp.), and KNN distance. Top 5 attack accuracies are included in the Supplementary \ref{sub:results_metric}. The best results are in \textbf{bold}.
By alleviating both these major problems in MI algorithms, we achieve new SOTA MI performance (face.evoLve: 81.40\% $\rightarrow$ \textbf{93.20\%}).}
\centering
\resizebox{0.9\columnwidth}{!}{%
\begin{tabular}{lccc}
\toprule
Method & \multicolumn{1}{l}{Attack Acc $\uparrow$} & \multicolumn{1}{l}{Imp. $\uparrow$} & \multicolumn{1}{l}{KNN Dist $\downarrow$} \\ \midrule

\multicolumn{4}{c}{\textbf{CelebA/CelebA/IR152}} \\ \midrule
KEDMI & 80.53 $\pm$ 3.86 & - & 1247.28 \\
 + LOM (Ours) & 92.47 $\pm$ 1.41 & 11.94 & 1168.55\\
+ MA (Ours) & 84.73 $\pm$ 3.76 & 4.20 & 1220.23\\
 + LOMMA (Ours) & \textbf{92.93 $\pm$ 1.15} & \textbf{12.40} & \textbf{1138.62}\\ \hdashline
GMI & 30.60 $\pm$ 6.54 & - & 1609.29\\
 + LOM (Ours) & 78.53 $\pm$ 3.41 & 47.93 & 1289.62\\
+ MA (Ours) & 61.20 $\pm$ 4.34 & 30.60 & 1389.99\\
 + LOMMA (Ours) & \textbf{82.40$\pm$ 4.37} & \textbf{51.80} & \textbf{1254.32}\\
\midrule

\multicolumn{4}{c}{\textbf{CelebA/CelebA/face.evoLve}} \\ \midrule
KEDMI & 81.40 $\pm$ 3.25 & - & 1248.32\\
 + LOM (Ours) & 92.53 $\pm$ 1.51 & 11.13 & 1183.76\\
+ MA (Ours) & 85.07 $\pm$ 2.71 & 3.67 & 1222.02\\
 + LOMMA (Ours) & \textbf{93.20 $\pm$ 0.85} & \textbf{11.80} & \textbf{1154.32}\\ \hdashline
GMI & 27.07 $\pm$ 6.72 & - & 1635.87\\
 + LOM (Ours) & 61.67 $\pm$ 4.92 & 34.60 & 1405.35\\
+ MA (Ours) & 74.13 $\pm$ 4.32 & 47.06 & 1352.25\\
 + LOMMA (Ours) & \textbf{82.33 $\pm$ 3.51} & \textbf{55.26} & \textbf{1257.50} \\\midrule

\multicolumn{4}{c}{\textbf{CelebA/CelebA/VGG16}} \\ \midrule
KEDMI & 74.00 $\pm$ 3.10 & - & 1289.88\\
 + LOM (Ours) & 89.07 $\pm$ 1.46 & 15.07 & 1218.46\\
+ MA (Ours) & 82.00 $\pm$ 3.85 & 8.00 & 1248.33\\
 + LOMMA (Ours) & \textbf{90.27 $\pm$ 1.36} & \textbf{16.27} & \textbf{1147.41} \\ \hdashline
GMI & 19.07 $\pm$ 4.47 & - & 1715.60\\
 + LOM (Ours) & 69.67 $\pm$ 4.80 & 50.60 & 1363.81\\
+ MA (Ours) & 51.73 $\pm$ 6.03 & 32.66 & 1467.68\\
 + LOMMA (Ours) & \textbf{77.60 $\pm$ 4.64} & \textbf{58.53} & \textbf{1296.26}\\ \bottomrule

\end{tabular}
}
\label{tab:celeba_celeba_results}
\vspace{-0.6cm}
\end{table}

\subsection{Experimental Results} 

\noindent
\textbf{Comparison with previous state-of-the-art.}
We use GMI \cite{zhang2020secret}, KEDMI \cite{chen2021knowledge}, and VMI \cite{wang2021variational} as SOTA MI baselines.
We reproduce all baseline results using official public implementations.
We report results for GMI and KEDMI for CelebA/ CelebA experiments in Table \ref{tab:celeba_celeba_results}.
We report VMI results for CelebA/ CelebA experiments in Table \ref{tab:vmi_results}. 
For each baseline setup, we report results for 3 variants: $\bullet$ \textit{LOM} (Logit Maximization, Sec. \ref{sub-sec:improved_formulation_of_identity_loss}), $\bullet$ \textit{MA} (Model Augmentation, Sec. \ref{sub-sec:mi_overfitting}), $\bullet$ \textit{LOMMA} (Logit Maximization + Model Augmentation). The details are as follows:
\begin{enumerate}
 \item + LOM (Ours): We replace existing identity loss, ${L}_{id}$ with our improved identity loss ${L}_{id}^{logit}$ (Sec. \ref{sub-sec:improved_formulation_of_identity_loss}).
 \item+ MA (Ours): We replace existing identity loss, ${L}_{id}$ with our proposed ${L}_{id}^{aug}$ (Sec. \ref{sub-sec:mi_overfitting}).
 \item+ LOMMA (Ours): We combine both ${L}_{id}^{logit}$ (Sec. \ref{sub-sec:improved_formulation_of_identity_loss}) and ${L}_{id}^{aug}$ (Sec. \ref{sub-sec:mi_overfitting}) for model inversion.
\end{enumerate}

As one can clearly observe from Table \ref{tab:celeba_celeba_results} and Table \ref{tab:vmi_results}, our proposed methods yield significant improvement in MI attack accuracy in \textit{all experiment setups} showing the efficacy of our proposed methods.
Further, by combining both our proposed methods, we significantly boost attack accuracy.
The KNN results also clearly show that our proposed methods are able to reconstruct data close to the private training data compared to existing SOTA MI algorithms.
Particularly, we improve the KEDMI baseline \cite{chen2021knowledge} attack accuracy by 12.4\% under IR152 target classifier. We show private training data and reconstructed samples for KEDMI \cite{chen2021knowledge} under IR152 target model including all 3 variants in Fig. \ref{fig:kedmi_samples}.
We remark that in the standard CelebA benchmark, our method boosts attack accuracy significantly thereby achieving more than 90\% attack accuracy (Table \ref{tab:celeba_celeba_results}) for the first time in contemporary MI literature.
We also include CIFAR-10, MNIST and additional results in Supplementary \ref{sub:results_cifar_minist}.

\begin{table}
\caption{We follow exact the experiment setup of \cite{wang2021variational} for VMI experiments. Specifically, we use StyleGAN\cite{Karras2020ada} and Flow model \cite{kingma2018glow} to learn the distribution of $z$. The best results are in \textbf{bold}. Following exact experiment setups in \cite{wang2021variational}, here $\mathcal{D}_{priv}$ = CelebA, $\mathcal{D}_{pub}$ = CelebA, target model = ResNet-34, evaluation model = IR-SE50.
We report top 1 attack accuracies, the improvement compared to the SOTA MI (Imp.), and KNN distance (KNN Dist). The top 5 attack accuracies are included in the Supplementary. The best results are in 
\textbf{bold}.
By alleviating both these major problems in MI algorithms, we improve the attack accuracy by 14.94\% (59.96\% $\rightarrow$ \textbf{74.90\%}).
}
\centering
\resizebox{0.88\columnwidth}{!}{%
\begin{tabular}{lccc}
\toprule
Method & \multicolumn{1}{l}{Attack Acc $\uparrow$} & \multicolumn{1}{l}{Imp. $\uparrow$} & \multicolumn{1}{l}{KNN Dist $\downarrow$} \\ 
\midrule
\multicolumn{4}{c}{\textbf{CelebA/CelebA/ResNet-34}} \\ \midrule
VMI & 59.96 $\pm$ 0.27 & - & 1.144 \\
+ LOM (Ours) & 68.34 $\pm$ 0.36 & 8.38 & 1.131 \\
+ MA (Ours) & 64.16 $\pm$ 0.27 & 4.20 & 1.140\\
+ LOMMA (Ours) & \textbf{ 74.90} $\pm$ 0.34 & \textbf{14.94} & \textbf{1.109} \\ \hline
\end{tabular}
}
\label{tab:vmi_results}
\vspace{-0.2cm}
\end{table}

\vspace{0.1cm}
\noindent
\textbf{Cross-dataset.}  
Following \cite{chen2021knowledge}, we conduct a series of experiments to study the effect of distribution shift between public and private data on attack performance and KNN distance.
We use FFHQ \cite{karras2019style} as the public dataset.
In particular, we use FFHQ as public data for CelebA experiments. We train GAN models and three model augmentations using the public data. 
We remark that such setups closely replicate real-world MI attack scenario.
We report top 1 accuracy and KNN distance for IR152, face.evoLve, and VGG16 target classifiers in Table \ref{tab:ffhq}. 
It is well known that baseline attack performances will degrade due to distribution shift between public and private data \cite{chen2021knowledge}. 
But we remark that our proposed methods consistently improves the baseline SOTA attack performances. 
\ie Our method boosts the attack accuracy of IR152 target model from 52.87\% $\rightarrow$ 77.27\%.

\vspace{0.15cm}
\noindent
\textbf{MI under SOTA defense models.}
We further evaluate our method on SOTA MI defense models provided by BiDO-HSIC \cite{peng2022bilateral}. 
Specifically, we use the exact GAN and defense models provided by BiDO-HSIC which are trained on CelebA dataset. 
We then transfer knowledge from the defense model to $M_{aug} = \{$Efficientnet-B0, Efficientnet-B1, Efficientnet-B2$\}$ using $D_{pub}$. 
Results using GMI and KEDMI are shown in Table \ref{tab:celeba_celeba_MI_defenses}.
We observe that SOTA defense BiDO-HSIC is rather ineffective for our proposed MI.

\begin{table}[!t]
\caption{ 
Results for
SOTA defense model BiDO-HSIC \cite{peng2022bilateral}: Following exact experiment setups in BiDO-HSIC, $\mathcal{D}_{priv}$ = CelebA, $\mathcal{D}_{pub}$ = CelebA, evaluation model = face.evoLve,
target model = BiDO-HSIC.
We report top 1 attack accuracies (Attack Acc.), and KNN distance (KNN Dist).
}
\vspace{-0.2cm}
\centering
\begin{adjustbox}{width=0.99\columnwidth,center}
\begin{tabular}{lcc|cc}
\hline
\multirow{2}{*}{Method} & \multicolumn{2}{c|}{GMI} & \multicolumn{2}{c}{KEDMI} \\\cline{2-5} 
 & Attack Acc $\uparrow$ & KNN Dist $\downarrow$ & Attack Acc $\uparrow$ & KNN Dist $\downarrow$ \\ \hline
 No Def. & 19.07 $\pm$ 4.47 & 1715.60 & 74.00 $\pm$ 3.10 & 1289.88 \\ \hline
Def. Model & 5.20 $\pm$ 2.75 & 1962.58 & 42.80 $\pm$ 5.02 & 1469.75 \\
 + LOM (Ours)& 55.80 $\pm$ 3.64 & 1397.05 & 64.33 $\pm$ 1.82 & 1360.57 \\
+ MA (Ours) & 23.93 $\pm$ 5.50 & 1634.84 & 49.27 $\pm$ 4.02 & 1413.81 \\
+ LOMMA (Ours) & \textbf{62.13} $\pm$ 4.04 & \textbf{1358.54} & \textbf{70.47} $\pm$ 2.36 & \textbf{1293.25} \\\hline
\end{tabular}

\end{adjustbox}
\vspace{-0.5cm}
\label{tab:celeba_celeba_MI_defenses}
\end{table}

\begin{table}
\caption{
We report the results for KEDMI and GMI for IR152, face.evoLve and VGG16 target model. Here $\mathcal{D}_{priv}$ = CelebA, $\mathcal{D}_{pub}$ = FFHQ, evaluation model = face.evoLve.
We report top 1 accuracies, the improvement compared to the SOTA MI (Imp.), and KNN distance. Top 5 attack accuracies are included in the Supplementary. The best results are in \textbf{bold}.
By alleviating both these major problems in MI algorithms, we improve the attack accuracy 24.40\% (IR152: 52.87\% $\rightarrow$ \textbf{77.27\%}).
}
\centering
\resizebox{0.89\columnwidth}{!}{%
\begin{tabular}{lccc}
\toprule
Method & \multicolumn{1}{l}{Attack Acc $\uparrow$} & \multicolumn{1}{l}{Imp. $\uparrow$} & \multicolumn{1}{l}{KNN Dist $\downarrow$} \\ \midrule

\multicolumn{4}{c}{\textbf{CelebA/FFHQ/IR152}} \\ \midrule
KEDMI & 52.87 $\pm$ 4.96 & - & 1418.83\\
+ LOM (Ours) & 67.73 $\pm$ 2.29 & 14.86 & 1325.28\\
+ MA (Ours) & 64.13 $\pm$ 4.49 & 11.26 & 1373.42\\
+ LOMMA (Ours) & \textbf{77.27 $\pm$ 2.01} & \textbf{24.40} & \textbf{1292.80} \\ \hdashline
GMI & 17.20 $\pm$ 5.31 & - & 1701.76\\
+ LOM (Ours) & 56.00 $\pm$ 5.20 & 38.80 & 1427.59\\
+ MA (Ours) & 50.80 $\pm$ 6.89 & 33.60 & 1462.92\\
+ LOMMA (Ours) & \textbf{72.00 $\pm$ 6.62} & \textbf{54.80} & \textbf{1338.35} \\ \hline

\multicolumn{4}{c}{\textbf{CelebA/FFHQ/face.evoLve}} \\ \hline
KEDMI & 51.87 $\pm$ 3.88 & - & 1440.19\\
+ LOM (Ours) & 69.73 $\pm$ 2.47 & 17.86 & 1379.73\\
+ MA (Ours) & 65.73 $\pm$ 3.51 & 13.86 & 1379.09\\
+ LOMMA (Ours) & \textbf{73.20 $\pm$ 2.24} & \textbf{21.33} & \textbf{1321.00}\\ \hdashline
GMI & 14.27 $\pm$ 4.42 & - & 1744.47\\
+ LOM (Ours) & 47.93 $\pm$ 4.87 & 33.66 & 1498.19\\
+ MA (Ours) & 46.07 $\pm$ 4.88 & 31.80 & 1500.10\\
+ LOMMA (Ours) & \textbf{64.33 $\pm$ 4.69} & \textbf{50.06} & \textbf{1386.33}\\ \hline

\multicolumn{4}{c}{\textbf{CelebA/FFHQ/VGG16}} \\ \hline
KEDMI & 41.27 $\pm$ 3.50 & - & 1490.09\\
+ LOM (Ours) & 55.07 $\pm$ 1.88 & 13.80 & 1438.72\\
+ MA (Ours) & 52.07 $\pm$ 2.92 & 10.80 & 1428.77\\
+ LOMMA (Ours) & \textbf{62.67 $\pm$ 2.29} & \textbf{21.40} & \textbf{1366.94}\\ \hdashline
GMI & 10.93 $\pm$ 3.47 & - & 1766.27\\
+ LOM (Ours) & 44.40 $\pm$ 5.96 & 33.47 & 1508.84\\
+ MA (Ours) & 34.93 $\pm$ 4.52 & 24.00 & 1547.93\\
+ LOMMA (Ours) & \textbf{58.73 $\pm$ 6.18} & \textbf{47.80} & \textbf{1415.06}\\ \hline

\end{tabular}
}
\label{tab:ffhq}
\vspace{-0.5cm}
\end{table}

\section{Discussion}

\noindent
\textbf{Conclusion.}
We revisit SOTA MI and study two issues pertaining to all SOTA MI approaches.
First, we 
analyze existing identity loss in SOTA and argue that it is sub-optimal for MI.
We propose a new logit based identity loss that aligns better with the goal of MI.
Second, we formalize the concept of MI overfitting and show that it has a considerable impact even in SOTA.
Inspired by   conventional data augmentation, we 
propose model augmentation to alleviate MI overfitting.
Extensive experiments demonstrate that our solutions can improve SOTA
significantly, achieving for the first time over 90\% attack
accuracy under the standard benchmark.
Our findings highlight rising threats based on MI and
 prompt serious consideration on privacy of machine learning.

\vspace{0.1cm}
\noindent
{\bf Limitations and Ethical Concerns.}
We follow  previous work in experimental setups. 
The scale of our experiments is comparable to previous works. 
Furthermore, extension of our methods for blackbox/ label-only attacks can be considered in future.
While our 
 improved MI methods could have 
negative societal impacts if it is used by malicious users, our work contributes to increased 
awareness about privacy attacks on DNNs.

\vspace{0.05cm}
\noindent
\textbf{Acknowledgements.}
This research is supported by the National Research Foundation, Singapore under its AI Singapore Programmes (AISG Award No.: AISG2-RP-2021-021; AISG Award No.: AISG2-TC-2022-007). 
This project is also supported by SUTD project PIE-SGP-AI-2018-01. 
We thank reviewers for their valuable comments.
We also thank Loo Yi and Kelly Kuo for helpful discussion.

{\small
\bibliographystyle{ieee_fullname}
\bibliography{ref}

\begin{thebibliography}{10}\itemsep=-1pt

\bibitem{abdollahzadeh2021revisit}
Milad Abdollahzadeh, Touba Malekzadeh, and Ngai-Man Cheung.
\newblock Revisit multimodal meta-learning through the lens of multi-task
  learning.
\newblock {\em Advances in Neural Information Processing Systems},
  34:14632--14644, 2021.

\bibitem{beaulieu2019privacy}
Brett~K Beaulieu-Jones, Zhiwei~Steven Wu, Chris Williams, Ran Lee, Sanjeev~P
  Bhavnani, James~Brian Byrd, and Casey~S Greene.
\newblock Privacy-preserving generative deep neural networks support clinical
  data sharing.
\newblock {\em Circulation: Cardiovascular Quality and Outcomes},
  12(7):e005122, 2019.

\bibitem{chabanne2017privacy}
Herv{\'e} Chabanne, Amaury De~Wargny, Jonathan Milgram, Constance Morel, and
  Emmanuel Prouff.
\newblock Privacy-preserving classification on deep neural network.
\newblock {\em Cryptology ePrint Archive}, 2017.

\bibitem{Chandrasegaran_2022_ECCV}
Keshigeyan Chandrasegaran, Ngoc-Trung Tran, Alexander Binder, and Ngai-Man
  Cheung.
\newblock {Discovering Transferable Forensic Features for CNN-generated Images
  Detection}.
\newblock In {\em Proceedings of the European Conference on Computer Vision
  (ECCV)}, Oct 2022.

\bibitem{Chandrasegaran_2021_CVPR}
Keshigeyan Chandrasegaran, Ngoc-Trung Tran, and Ngai-Man Cheung.
\newblock {A Closer Look at Fourier Spectrum Discrepancies for CNN-Generated
  Images Detection}.
\newblock In {\em Proceedings of the IEEE/CVF Conference on Computer Vision and
  Pattern Recognition (CVPR)}, pages 7200--7209, June 2021.

\bibitem{pmlr-v162-chandrasegaran22a}
Keshigeyan Chandrasegaran, Ngoc-Trung Tran, Yunqing Zhao, and Ngai-Man Cheung.
\newblock Revisiting label smoothing and knowledge distillation compatibility:
  What was missing?
\newblock In {\em Proceedings of the 39th International Conference on Machine
  Learning}, volume 162 of {\em Proceedings of Machine Learning Research},
  pages 2890--2916. PMLR, 17-23 Jul 2022.

\bibitem{chen2021knowledge}
Si Chen, Mostafa Kahla, Ruoxi Jia, and Guo-Jun Qi.
\newblock Knowledge-enriched distributional model inversion attacks.
\newblock In {\em Proceedings of the IEEE/CVF international conference on
  computer vision}, pages 16178--16187, 2021.

\bibitem{yu_know_you_at_one_glance}
Yu Cheng, Jian Zhao, Zhecan Wang, Yan Xu, Karlekar Jayashree, Shengmei Shen,
  and Jiashi Feng.
\newblock Know you at one glance: A compact vector representation for low-shot
  learning.
\newblock In {\em 2017 IEEE International Conference on Computer Vision
  Workshops (ICCVW)}, pages 1924--1932, 2017.

\bibitem{cheng2017know}
Yu Cheng, Jian Zhao, Zhecan Wang, Yan Xu, Karlekar Jayashree, Shengmei Shen,
  and Jiashi Feng.
\newblock Know you at one glance: A compact vector representation for low-shot
  learning.
\newblock In {\em Proceedings of the IEEE International Conference on Computer
  Vision Workshops}, pages 1924--1932, 2017.

\bibitem{choquette2021label}
Christopher~A Choquette-Choo, Florian Tramer, Nicholas Carlini, and Nicolas
  Papernot.
\newblock Label-only membership inference attacks.
\newblock In {\em International conference on machine learning}, pages
  1964--1974. PMLR, 2021.

\bibitem{cohen2017emnist}
Gregory Cohen, Saeed Afshar, Jonathan Tapson, and Andre Van~Schaik.
\newblock Emnist: Extending mnist to handwritten letters.
\newblock In {\em 2017 International Joint Conference on Neural Networks
  (IJCNN)}, pages 2921--2926. IEEE, 2017.

\bibitem{deng2019arcface}
Jiankang Deng, Jia Guo, Niannan Xue, and Stefanos Zafeiriou.
\newblock Arcface: Additive angular margin loss for deep face recognition.
\newblock In {\em Proceedings of the IEEE/CVF conference on computer vision and
  pattern recognition}, pages 4690--4699, 2019.

\bibitem{fredrikson2015model}
Matt Fredrikson, Somesh Jha, and Thomas Ristenpart.
\newblock Model inversion attacks that exploit confidence information and basic
  countermeasures.
\newblock In {\em Proceedings of the 22nd ACM SIGSAC conference on computer and
  communications security}, pages 1322--1333, 2015.

\bibitem{fredrikson2014privacy}
Matthew Fredrikson, Eric Lantz, Somesh Jha, Simon Lin, David Page, and Thomas
  Ristenpart.
\newblock Privacy in pharmacogenetics: An $\{$End-to-End$\}$ case study of
  personalized warfarin dosing.
\newblock In {\em 23rd USENIX Security Symposium (USENIX Security 14)}, pages
  17--32, 2014.

\bibitem{goodfellow2020generative}
Ian Goodfellow, Jean Pouget-Abadie, Mehdi Mirza, Bing Xu, David Warde-Farley,
  Sherjil Ozair, Aaron Courville, and Yoshua Bengio.
\newblock Generative adversarial networks.
\newblock {\em Communications of the ACM}, 63(11):139--144, 2020.

\bibitem{he2020momentum}
Kaiming He, Haoqi Fan, Yuxin Wu, Saining Xie, and Ross Girshick.
\newblock {Momentum contrast for unsupervised visual representation learning}.
\newblock In {\em Proceedings of the IEEE/CVF conference on computer vision and
  pattern recognition}, pages 9729--9738, 2020.

\bibitem{he2016deep}
Kaiming He, Xiangyu Zhang, Shaoqing Ren, and Jian Sun.
\newblock Deep residual learning for image recognition.
\newblock In {\em Proceedings of the IEEE conference on computer vision and
  pattern recognition}, pages 770--778, 2016.

\bibitem{heusel2017gans}
Martin Heusel, Hubert Ramsauer, Thomas Unterthiner, Bernhard Nessler, and Sepp
  Hochreiter.
\newblock Gans trained by a two time-scale update rule converge to a local nash
  equilibrium.
\newblock {\em Advances in neural information processing systems}, 30, 2017.

\bibitem{hinton2015distilling}
Geoffrey Hinton, Oriol Vinyals, and Jeff Dean.
\newblock Distilling the knowledge in a neural network, 2015.

\bibitem{howard2019searching}
Andrew Howard, Mark Sandler, Grace Chu, Liang-Chieh Chen, Bo Chen, Mingxing
  Tan, Weijun Wang, Yukun Zhu, Ruoming Pang, Vijay Vasudevan, et~al.
\newblock Searching for mobilenetv3.
\newblock In {\em Proceedings of the IEEE/CVF international conference on
  computer vision}, pages 1314--1324, 2019.

\bibitem{huang2017densely}
Gao Huang, Zhuang Liu, Laurens Van Der~Maaten, and Kilian~Q Weinberger.
\newblock Densely connected convolutional networks.
\newblock In {\em Proceedings of the IEEE conference on computer vision and
  pattern recognition}, pages 4700--4708, 2017.

\bibitem{kahla2022label}
Mostafa Kahla, Si Chen, Hoang~Anh Just, and Ruoxi Jia.
\newblock Label-only model inversion attacks via boundary repulsion.
\newblock In {\em Proceedings of the IEEE/CVF Conference on Computer Vision and
  Pattern Recognition}, pages 15045--15053, 2022.

\bibitem{Karras2020ada}
Tero Karras, Miika Aittala, Janne Hellsten, Samuli Laine, Jaakko Lehtinen, and
  Timo Aila.
\newblock Training generative adversarial networks with limited data.
\newblock In {\em Proc. NeurIPS}, 2020.

\bibitem{karras2019style}
Tero Karras, Samuli Laine, and Timo Aila.
\newblock A style-based generator architecture for generative adversarial
  networks.
\newblock In {\em Proceedings of the IEEE/CVF conference on computer vision and
  pattern recognition}, pages 4401--4410, 2019.

\bibitem{khosla2020supervised}
Prannay Khosla, Piotr Teterwak, Chen Wang, Aaron Sarna, Yonglong Tian, Phillip
  Isola, Aaron Maschinot, Ce Liu, and Dilip Krishnan.
\newblock Supervised contrastive learning.
\newblock {\em Advances in neural information processing systems},
  33:18661--18673, 2020.

\bibitem{kingma2018glow}
Durk~P Kingma and Prafulla Dhariwal.
\newblock Glow: Generative flow with invertible 1x1 convolutions.
\newblock {\em Advances in neural information processing systems}, 31, 2018.

\bibitem{koh2023grounding}
Jing~Yu Koh, Ruslan Salakhutdinov, and Daniel Fried.
\newblock Grounding language models to images for multimodal generation.
\newblock {\em arXiv preprint arXiv:2301.13823}, 2023.

\bibitem{krizhevsky2010cifar}
Alex Krizhevsky, Vinod Nair, and Geoffrey Hinton.
\newblock Cifar-10 (canadian institute for advanced research).
\newblock {\em URL http://www. cs. toronto. edu/kriz/cifar. html}, 5(4):1,
  2010.

\bibitem{alexnet}
Alex Krizhevsky, Ilya Sutskever, and Geoffrey~E Hinton.
\newblock Imagenet classification with deep convolutional neural networks.
\newblock In F. Pereira, C.J. Burges, L. Bottou, and K.Q. Weinberger, editors,
  {\em Advances in Neural Information Processing Systems}, volume~25. Curran
  Associates, Inc., 2012.

\bibitem{lecun1998gradient}
Yann LeCun, L{\'e}on Bottou, Yoshua Bengio, and Patrick Haffner.
\newblock Gradient-based learning applied to document recognition.
\newblock {\em Proceedings of the IEEE}, 86(11):2278--2324, 1998.

\bibitem{lee2022privacy}
Joon-Woo Lee, HyungChul Kang, Yongwoo Lee, Woosuk Choi, Jieun Eom, Maxim
  Deryabin, Eunsang Lee, Junghyun Lee, Donghoon Yoo, Young-Sik Kim, et~al.
\newblock Privacy-preserving machine learning with fully homomorphic encryption
  for deep neural network.
\newblock {\em IEEE Access}, 10:30039--30054, 2022.

\bibitem{lim2018doping}
Swee~Kiat Lim, Yi Loo, Ngoc~Trung Tran, Ngai~Man Cheung, Gemma Roig, and Yuval
  Elovici.
\newblock Doping: Generative data augmentation for unsupervised anomaly
  detection with gan.
\newblock In {\em 18th IEEE International Conference on Data Mining, ICDM
  2018}, pages 1122--1127. Institute of Electrical and Electronics Engineers
  Inc., 2018.

\bibitem{liu2015deep}
Ziwei Liu, Ping Luo, Xiaogang Wang, and Xiaoou Tang.
\newblock Deep learning face attributes in the wild.
\newblock In {\em Proceedings of the IEEE international conference on computer
  vision}, pages 3730--3738, 2015.

\bibitem{lukasik20a}
Michal Lukasik, Srinadh Bhojanapalli, Aditya Menon, and Sanjiv Kumar.
\newblock Does label smoothing mitigate label noise?
\newblock In Hal~Daumé III and Aarti Singh, editors, {\em Proceedings of the
  37th International Conference on Machine Learning}, volume 119 of {\em
  Proceedings of Machine Learning Research}, pages 6448--6458. PMLR, 13--18 Jul
  2020.

\bibitem{muller}
Rafael M\"{u}ller, Simon Kornblith, and Geoffrey~E Hinton.
\newblock When does label smoothing help?
\newblock In H. Wallach, H. Larochelle, A. Beygelzimer, F. d\textquotesingle
  Alch\'{e}-Buc, E. Fox, and R. Garnett, editors, {\em Advances in Neural
  Information Processing Systems}, volume~32. Curran Associates, Inc., 2019.

\bibitem{peng2022bilateral}
Xiong Peng, Feng Liu, Jingfen Zhang, Long Lan, Junjie Ye, Tongliang Liu, and Bo
  Han.
\newblock Bilateral dependency optimization: Defending against model-inversion
  attacks.
\newblock {\em KDD}, 2022.

\bibitem{radford2021learning}
Alec Radford, Jong~Wook Kim, Chris Hallacy, Aditya Ramesh, Gabriel Goh,
  Sandhini Agarwal, Girish Sastry, Amanda Askell, Pamela Mishkin, Jack Clark,
  et~al.
\newblock Learning transferable visual models from natural language
  supervision.
\newblock In {\em International conference on machine learning}, pages
  8748--8763. PMLR, 2021.

\bibitem{santos2022avoiding}
Claudio Filipi Gon{\c{c}}alves~Dos Santos and Jo{\~a}o~Paulo Papa.
\newblock Avoiding overfitting: A survey on regularization methods for
  convolutional neural networks.
\newblock {\em ACM Computing Surveys (CSUR)}, 54(10s):1--25, 2022.

\bibitem{shen2021}
Zhiqiang Shen, Zechun Liu, Dejia Xu, Zitian Chen, Kwang-Ting Cheng, and Marios
  Savvides.
\newblock Is label smoothing truly incompatible with knowledge distillation: An
  empirical study.
\newblock In {\em International Conference on Learning Representations}, 2021.

\bibitem{simonyan2014very}
Karen Simonyan and Andrew Zisserman.
\newblock Very deep convolutional networks for large-scale image recognition.
\newblock {\em arXiv preprint arXiv:1409.1556}, 2014.

\bibitem{sirichotedumrong2021gan}
Warit Sirichotedumrong and Hitoshi Kiya.
\newblock A gan-based image transformation scheme for privacy-preserving deep
  neural networks.
\newblock In {\em 2020 28th European Signal Processing Conference (EUSIPCO)},
  pages 745--749. IEEE, 2021.

\bibitem{subbanna2021analysis}
Nagesh Subbanna, Matthias Wilms, Anup Tuladhar, and Nils~D Forkert.
\newblock An analysis of the vulnerability of two common deep learning-based
  medical image segmentation techniques to model inversion attacks.
\newblock {\em Sensors}, 21(11):3874, 2021.

\bibitem{tan2019efficientnet}
Mingxing Tan and Quoc Le.
\newblock Efficientnet: Rethinking model scaling for convolutional neural
  networks.
\newblock In {\em International conference on machine learning}, pages
  6105--6114. PMLR, 2019.

\bibitem{teo2022fair}
Christopher~TH Teo, Milad Abdollahzadeh, and Ngai-Man Cheung.
\newblock Fair generative models via transfer learning.
\newblock {\em arXiv preprint arXiv:2212.00926}, 2022.

\bibitem{tran2021data}
Ngoc-Trung Tran, Viet-Hung Tran, Ngoc-Bao Nguyen, Trung-Kien Nguyen, and
  Ngai-Man Cheung.
\newblock On data augmentation for gan training.
\newblock {\em IEEE Transactions on Image Processing}, 30:1882--1897, 2021.

\bibitem{tran2021data_aug_gan}
Ngoc-Trung Tran, Viet-Hung Tran, Ngoc-Bao Nguyen, Trung-Kien Nguyen, and
  Ngai-Man Cheung.
\newblock On data augmentation for gan training.
\newblock {\em IEEE Transactions on Image Processing}, 30:1882--1897, 2021.

\bibitem{wang2021variational}
Kuan-Chieh Wang, Yan Fu, Ke Li, Ashish Khisti, Richard Zemel, and Alireza
  Makhzani.
\newblock Variational model inversion attacks.
\newblock {\em Advances in Neural Information Processing Systems},
  34:9706--9719, 2021.

\bibitem{yang2019adversarial}
Ziqi Yang, Ee-Chien Chang, and Zhenkai Liang.
\newblock Adversarial neural network inversion via auxiliary knowledge
  alignment.
\newblock {\em arXiv preprint arXiv:1902.08552}, 2019.

\bibitem{yang2019neural}
Ziqi Yang, Jiyi Zhang, Ee-Chien Chang, and Zhenkai Liang.
\newblock Neural network inversion in adversarial setting via background
  knowledge alignment.
\newblock In {\em Proceedings of the 2019 ACM SIGSAC Conference on Computer and
  Communications Security}, pages 225--240, 2019.

\bibitem{yu2022understanding}
Chaojian Yu, Bo Han, Li Shen, Jun Yu, Chen Gong, Mingming Gong, and Tongliang
  Liu.
\newblock Understanding robust overfitting of adversarial training and beyond.
\newblock In {\em International Conference on Machine Learning}, pages
  25595--25610. PMLR, 2022.

\bibitem{zhang2021cross}
Han Zhang, Jing~Yu Koh, Jason Baldridge, Honglak Lee, and Yinfei Yang.
\newblock Cross-modal contrastive learning for text-to-image generation.
\newblock In {\em Proceedings of the IEEE/CVF conference on computer vision and
  pattern recognition}, pages 833--842, 2021.

\bibitem{zhang2020secret}
Yuheng Zhang, Ruoxi Jia, Hengzhi Pei, Wenxiao Wang, Bo Li, and Dawn Song.
\newblock The secret revealer: Generative model-inversion attacks against deep
  neural networks.
\newblock In {\em Proceedings of the IEEE/CVF Conference on Computer Vision and
  Pattern Recognition}, pages 253--261, 2020.

\bibitem{zhao2021exploiting}
Xuejun Zhao, Wencan Zhang, Xiaokui Xiao, and Brian Lim.
\newblock Exploiting explanations for model inversion attacks.
\newblock In {\em Proceedings of the IEEE/CVF International Conference on
  Computer Vision}, pages 682--692, 2021.

\bibitem{zhao2022fewshot}
Yunqing Zhao, Keshigeyan Chandrasegaran, Milad Abdollahzadeh, and Ngai-Man
  Cheung.
\newblock Few-shot image generation via adaptation-aware kernel modulation.
\newblock In {\em Advances in Neural Information Processing Systems}, 2022.

\end{thebibliography}
}

\newpage
\appendix
\unhidefromtoc

{\centering \large \textbf{Supplementary Materials}}

\noindent
In this supplementary material, we provide additional experiments, analysis, ablation study, and reproducibility details 
to support our findings.
We provide Pytorch code, demo and pre-trained models (target models/ evaluation models/ augmented models) at: \textcolor{magenta}{\url{https://ngoc-nguyen-0.github.io/re-thinking_model_inversion_attacks/}}.
 
\thispagestyle{empty}
\tableofcontents

\renewcommand\thefigure{\thesection.\arabic{figure}}
\renewcommand\theHfigure{\thesection.\arabic{figure}}
\renewcommand\thetable{\thesection.\arabic{table}}  
\renewcommand\theHtable{\thesection.\arabic{table}}  

\setcounter{figure}{0} 
\setcounter{table}{0}

\section{Additional experimental results}
\label{supp-sec:additional_results}
In this section, we provide additional experimental results that are not included in the main paper.
More specifically, first, we evaluate the effect of the proposed method on improving SOTA approaches in new tasks including image classification and digit classification. Then, we use alternative metrics for evaluating SOTA MI approaches with and without proposed improvements on identity loss $L_{id}$. The additional experimental results in this section further support effectiveness of the proposed approach on improving MI attacks.

\subsection{Experimental results on CIFAR-10 and MNIST}
\label{sub:results_cifar_minist}
In Sec. 4.2 of the main paper, we mostly focus on the face recognition task (on the CelebA dataset) and show that the proposed method significantly improves SOTA approaches by increasing {\bf Attack Acc} (inference accuracy on reconstructed samples by an evaluation model; see Sec. 4.1. of the main paper) and decreasing {\bf KNN Dist} (distance between the reconstructed samples of a specific class/id and corresponding data in the private dataset $\mathcal{D}_{priv}$; see Sec. 4.1).

\begin{table}[h]
\caption{
We report top 1 accuracies, the improvement compared to the SOTA MI (Imp.), and KNN distance for two experiment setups. Following exact experiment setups in \cite{chen2021knowledge}. For CIFAR-10 experiments, 
$\mathcal{D}_{priv}$ = CIFAR-10, $\mathcal{D}_{pub}$ = CIFAR-10, $M_t$ = VGG16, evaluation model = ResNet-18. For MNIST experiments, $\mathcal{D}_{priv}$ = MNIST, $\mathcal{D}_{pub}$ = MNIST, $M_t$ = CNN(Conv3), evaluation model = CNN(Conv5). The best results are in \textbf{bold}. 
}

\centering
\resizebox{0.9\columnwidth}{!}{%
\begin{tabular}{lccc}
\toprule
Method & \multicolumn{1}{l}{Attack Acc $\uparrow$} & \multicolumn{1}{l}{Imp. $\uparrow$} & \multicolumn{1}{l}{KNN Dist $\downarrow$} \\ \midrule

\multicolumn{4}{c}{\textbf{CIFAR-10/CIFAR-10/VGG16}} \\ \midrule
KEDMI & 95.2 $\pm$ 7.96 & - & 78.24 \\
+ LOM (Ours) & 100 $\pm$ 0 & 4.80 & \textbf{52.12}\\
+ MA (Ours) & 100 $\pm$ 0 & 4.80 & 53.17\\
+ LOMMA (Ours) & \textbf{100 $\pm$ 0} & \textbf{4.80} &  63.41\\ \hdashline
GMI & 43.20 $\pm$ 19.80 & - & 96.11 \\
+ LOM (Ours) & 80.80 $\pm$ 14.65 & 37.60 & \textbf{70.47} \\
+ MA (Ours) & 80.00 $\pm$ 18.01 & 36.80 & 93.46\\
+ LOMMA (Ours) & \textbf{95.20 $\pm$ 7.96} & \textbf{52.00} & 80.30\\
\midrule

\multicolumn{4}{c}{\textbf{MNIST/MNIST/CNN(Conv3)}} \\ \midrule
KEDMI & 46.40 $\pm$ 14.65 & - & 120.99 \\
+ LOM (Ours) & 55.20 $\pm$ 8.94 & 8.80 & 100.18 \\
+ MA (Ours) & 75.20 $\pm$ 6.57 & 28.80 & 72.38 \\
+ LOMMA (Ours) & \textbf{100.00 $\pm$ 0.00} & \textbf{53.60} & \textbf{58.81} \\ \hdashline
GMI &  8.00 $\pm$ 1.52 & - & 126.61 \\
+ LOM (Ours) & 15.20 $\pm$ 15.12 & 7.20 & 161.90 \\
+ MA (Ours) & 66.40 $\pm$ 19.86 & 58.40 & \textbf{78.38} \\
+ LOMMA (Ours) & \textbf{80.80 $\pm$ 17.38} & \textbf{72.80} & 83.56 \\ \hline

\end{tabular}
}
\label{tab:cifar_mnist_results}
\end{table}

\begin{table}[h]
\caption{We follow exact the experiment setup of \cite{wang2021variational} for the VMI experiments. Specifically, we use DCGAN and Flow model to learn the distribution of $\bz$. 
}
\centering
\resizebox{0.9\columnwidth}{!}{%
\begin{tabular}{lccc}
\toprule
Method & \multicolumn{1}{l}{Attack Acc $\uparrow$} & \multicolumn{1}{l}{Imp. $\uparrow$} & \multicolumn{1}{l}{KNN Dist $\downarrow$} \\ 
\midrule
\multicolumn{4}{c}{\textbf{MNIST/EMNIST/ResNet-10}} \\ \midrule
VMI & 94.60 $\pm$ 0.13 & - & 68.53 \\
+ LOM (Ours) & 98.60 $\pm$ 0.09 & 4.00 & 88.13\\
+ MA (Ours) & 98.98 $\pm$ 0.02  & 4.38 & 58.81 \\
+ LOMMA (Ours) & \textbf{100.00} $\pm$ 0.00 & 5.40 & \textbf{52.62} \\ \bottomrule

\end{tabular}
}
\label{tab:vmi_mnist}
\end{table}

In this section, we provide results for other tasks. 
More specifically, as mentioned in Sec. 4.1, for GMI~\cite{zhang2020secret}, and KEDMI~\cite{chen2021knowledge}, following their own setup, we use digit classification task MNIST dataset, and object classification task on the CIFAR-10 dataset. For each task, Table \ref{tab:cifar_mnist_results} tabulates the performance of the SOTA approach together with three variants of our proposed approach: 
\begin{enumerate}
 \item + LOM (Ours): We replace existing identity loss, ${L}_{id}$ with our improved identity loss ${L}_{id}^{logit}$ (Sec. 3.1).
 \item+ MA (Ours): We replace existing identity loss, ${L}_{id}$ with our proposed ${L}_{id}^{aug}$ (Sec. 3.2).
 \item+ LOMMA (Ours): We combine both ${L}_{id}^{logit}$ and ${L}_{id}^{aug}$ for model inversion.
\end{enumerate}
As one can see, on average each of the proposed solutions drastically improves the SOTA approaches, and combining these two solutions works even better.

\begin{table}[h]
\caption{
We report the results for KEDMI and GMI for IR152, face.evoLve and VGG16 target model. Following exact experiment setups in \cite{chen2021knowledge}, here $\mathcal{D}_{priv}$ = CelebA, $\mathcal{D}_{pub}$ = CelebA, evaluation model = face.evoLve.
We report top-5 accuracies, the improvement compared to the SOTA MI (Imp.), and FID scores.
}

\centering
\resizebox{0.9\columnwidth}{!}{%
\begin{tabular}{lccc}
\toprule
Method & \multicolumn{1}{l}{Top-5 Attack Acc $\uparrow$} & \multicolumn{1}{l}{Imp. $\uparrow$} & \multicolumn{1}{l}{FID $\downarrow$} \\ \midrule

\multicolumn{4}{c}{\textbf{CelebA/CelebA/IR152}} \\ \midrule
KEDMI & 98.00 $\pm$ 1.96 & - & \textbf{28.06} \\
+ LOM (Ours) & \textbf{98.67} $\pm$ 0.00 &  \textbf{0.67} & 39.03 \\
+ MA (Ours) & 98.33 $\pm$ 1.19 & 0.33 & 28.38 \\
+ LOMMA (Ours) & \textbf{98.67} $\pm$ 0.37 & \textbf{0.67} & 36.78 \\ \hdashline
GMI & 55.67 $\pm$ 7.14 & - & 57.11 \\
+ LOM (Ours) & 93.00 $\pm$ 3.41 & 37.33 & 48.87 \\
+ MA (Ours) & 89.00 $\pm$ 4.10 & 33.33 & 45.24 \\
+ LOMMA (Ours) & \textbf{97.67 $\pm$ 2.41} & \textbf{42.00} & \textbf{45.02} \\ \midrule

\multicolumn{4}{c}{\textbf{CelebA/CelebA/face.evoLve}} \\ \midrule
KEDMI & 97.33 $\pm$ 1.73 & - &  \textbf{31.26} \\
+ LOM (Ours) &  \textbf{99.33} $\pm$ 0.18 &  \textbf{2.00} & 42.45 \\
+ MA (Ours) & 98.00 $\pm$ 0.94 & 0.67 & 32.08 \\
+ LOMMA (Ours) &  \textbf{99.33} $\pm$ 0.33 &  \textbf{2.00} & 38.69 \\ \hdashline
GMI & 45.33 $\pm$ 8.05 & - & 59.76 \\
+ LOM (Ours) & 84.33 $\pm$ 4.49 & 39.00 & 44.27 \\
+ MA (Ours) & 92.00 $\pm$ 2.25 & 46.67 & 51.15 \\
+ LOMMA (Ours) &  \textbf{93.67 $\pm$ 2.42} &  \textbf{48.33} &  \textbf{44.07} \\ \midrule

\multicolumn{4}{c}{\textbf{CelebA/CelebA/VGG16}} \\ \midrule
KEDMI & 93.33 $\pm$ 3.36 & - & 25.46 \\
+ LOM (Ours) &  \textbf{99.00 $\pm$ 0.18} &  \textbf{5.67} & 34.45 \\
+ MA (Ours) & 95.33 $\pm$ 1.60 & 2.00 &  \textbf{24.65} \\
+ LOMMA (Ours) & 98.00 $\pm$ 0.61 & 4.67 & 33.91 \\ \hdashline
GMI & 40.33 $\pm$ 4.74 & - & 58.03 \\
+ LOM (Ours) & 89.33 $\pm$ 2.73 & 49.00 & 46.40 \\
+ MA (Ours) & 81.33 $\pm$ 5.88 & 41.00 & 44.90 \\
+ LOMMA (Ours) &  \textbf{95.67 $\pm$ 2.16} &  \textbf{55.34} &  \textbf{43.21} \\ \bottomrule

\end{tabular}
}
\label{tab:top5_celeba}
\end{table}

Additionally, as mentioned in Sec. 4.1, for VMI~\cite{wang2021variational}, following the setup in ~\cite{wang2021variational}, we evaluate its performance for digit classification on MNIST, and improvement brought by the proposed method. Note that for a fair comparison, following VMI implementation in ~\cite{wang2021variational}, in this experiment we use EMNIST~\cite{cohen2017emnist}  as public dataset $\mathcal{D}_{pub}$ to acquire prior knowledge. Results are shown in Table \ref{tab:vmi_mnist} for three variants of our proposed method, which indicates better performance in terms of both attack accuracy (reaching 100\% attack accuracy) and decreasing KNN Distance.

\subsection{Experimental Results with Additional Metrics}
\label{sub:results_metric}
As mentioned in Sec. 4.1 of the main paper, Attack Acc and KNN Dist are common metrics used in literature to evaluate the MI attacks. In this section, we include results on two additional metrics namely: Top-5 Attack Acc 
and FID~\cite{heusel2017gans}. Results in Table \ref{tab:top5_celeba}, Table \ref{tab:top5_vmi}, and Table \ref{tab:top5_ffhq} show that the proposed method achieves better performance in terms of Top-5 Attack Acc, and FID value.

\begin{table}[h]
\caption{We report the results for VMI . Following exact experiment setups in \cite{wang2021variational}, here $\mathcal{D}_{priv}$ = CelebA, $\mathcal{D}_{pub}$ = CelebA, $M_t$ = ResNet-34, evaluation model = IR-SE50.
We report top-5 accuracies, the improvement compared to the SOTA MI (Imp.), and FID scores.}

\centering
\resizebox{0.9\columnwidth}{!}{%
\begin{tabular}{lccc}
\toprule
Method & \multicolumn{1}{l}{Top-5 Attack Acc $\uparrow$} & \multicolumn{1}{l}{Imp. $\uparrow$} & \multicolumn{1}{l}{FID $\downarrow$} \\ \midrule

\multicolumn{4}{c}{\textbf{CelebA/CelebA/ResNet-34}} \\ \midrule
VMI & 82.32 $\pm$ 0.21 & - & \textbf{16.82}   \\
+ LOM (Ours) &  86.56 $\pm$ 0.27 & 4.24 & 25.42 \\
+ MA (Ours) & 86.16 $\pm$ 0.19 & 3.84 & 17.60  \\
+ LOMMA (Ours) & \textbf{91.02 $\pm$ 0.22} & \textbf{8.70} & 23.56  \\ \bottomrule
\end{tabular}
}
\label{tab:top5_vmi}
\end{table}

\begin{table}[t]
\caption{We report the results for KEDMI and GMI for IR152, face.evoLve and VGG16 target model. Following exact experiment setups in \cite{chen2021knowledge}, here $\mathcal{D}_{priv}$ = CelebA, $\mathcal{D}_{pub}$ = FFHQ, evaluation model = face.evoLve.
We report top-5 accuracies, the improvement compared to the SOTA MI (Imp.), and FID scores.}

\centering
\resizebox{0.9\columnwidth}{!}{%
\begin{tabular}{lccc}
\toprule
Method & \multicolumn{1}{l}{Top-5 Attack Acc $\uparrow$} & \multicolumn{1}{l}{Imp. $\uparrow$} & \multicolumn{1}{l}{FID $\downarrow$} \\ \midrule

\multicolumn{4}{c}{\textbf{CelebA/FFHQ/IR152}} \\ \midrule
KEDMI & 85.33 $\pm$ 4.01 & - & 41.71 \\
+ LOM (Ours) & 88.67 $\pm$ 1.18 & 3.33 & 50.84 \\
+ MA (Ours) & 87.67 $\pm$ 2.28 & 2.33 & \textbf{39.88} \\
+ LOMMA (Ours) & \textbf{92.00 $\pm$ 0.57} & \textbf{6.60} & 45.67 \\ \hdashline
GMI & 36.33 $\pm$ 3.98 & - & 47.72 \\
+ LOM (Ours) & 80.33 $\pm$ 4.21 & 44.00 & 40.18 \\
+ MA (Ours) & 84.00 $\pm$ 5.35 & 47.67 & \textbf{35.41} \\
+ LOMMA (Ours) & \textbf{90.33 $\pm$ 3.16} & \textbf{54.00} & 37.58 \\ \midrule

\multicolumn{4}{c}{\textbf{CelebA/FFHQ/face.evoLve}} \\ \midrule
KEDMI & 80.67 $\pm$ 2.83 & - & 38.09 \\
+ LOM (Ours) & 91.33 $\pm$ 0.47 & 10.67 & 47.30 \\
+ MA (Ours) & 88.67 $\pm$ 2.44 & 8.00 & \textbf{35.94} \\
+ LOMMA (Ours) & \textbf{94.00 $\pm$ 0.68} & \textbf{13.33} & 47.51 \\ \hdashline
GMI & 33.33 $\pm$ 6.18 & - & 52.84 \\
+ LOM (Ours) & 74.67 $\pm$ 4.78 & 41.33 & 44.01 \\
+ MA (Ours) & 72.00 $\pm$ 4.64 & 38.67 & \textbf{35.58} \\
+ LOMMA (Ours) & \textbf{89.00 $\pm$ 2.73} & \textbf{55.67} & 40.03 \\ \midrule

\multicolumn{4}{c}{\textbf{CelebA/FFHQ/VGG16}} \\ \midrule
KEDMI & 74.00 $\pm$ 4.05 & - & 36.18 \\
+ LOM (Ours) & 81.67 $\pm$ 1.19 & 7.67 & 43.76 \\
+ MA (Ours) & 80.33 $\pm$ 3.27 & 6.33 & \textbf{35.02} \\
+ LOMMA (Ours) & \textbf{85.33 $\pm$ 1.98} & \textbf{11.33} & 40.26 \\ \hdashline
GMI & 25.67 $\pm$ 5.13 & - & 53.17 \\
+ LOM (Ours) & 70.67 $\pm$ 3.92 & 45.00 & 42.60 \\
+ MA (Ours) & 62.33 $\pm$ 5.36 & 36.67 & 36.04 \\
+ LOMMA (Ours) & \textbf{86.33 $\pm$ 5.17} & \textbf{60.67} & \textbf{35.59} \\ \bottomrule

\end{tabular}
}
\label{tab:top5_ffhq}
\end{table}

\setcounter{figure}{0} 
\setcounter{table}{0} 

\section{Ablation Study}
\label{supp-sec:ablation_study}

\subsection{\boldmath  Different number of augmented models $M_{aug}$}
\label{sub:no_Maug}

In Sec 3.2, we propose a model augmentation idea with augmented models $M_{aug}$. Here, we experiment using a different number of networks for $M_{aug}$. 
Table \ref{tab:no_M_aug} show that increasing the number of the augmented models will improve attack accuracy. We use 3 augmented models in our main result as this configuration achieves a good tradeoff in accuracy and computation.

\begin{table*}[t]
\caption{We report top-1 attack accuracies, the improvement compared to the SOTA MI (Imp.), and KNN distance for using different numbers $N_{aug}$ of network $M_{aug}$. Following exact experiment setups in \cite{chen2021knowledge}, here method = KEDMI, $\mathcal{D}_{priv}$ = CelebA, $\mathcal{D}_{pub}$ = CelebA, $M_t$ = IR152, evaluation model = face.evoLve. We select $M_{aug}$ from the set of 4 networks including EfficientNet-B0, EfficientNet-B1, EfficientNet-B2, EfficientNet-B3.  The number of network $M_{aug}$ increases from 0 (Baseline KEDMI) to 4. It shows that using more $M_{aug}$ improves the attack accuracy and KNN distance.}
\centering
\resizebox{0.9\width}{!}{%
\begin{tabular}{lclccc}
\toprule
Method & $N_{aug}$ & $M_{aug}$ & \multicolumn{1}{l}{Attack Acc $\uparrow$} & \multicolumn{1}{l}{Imp. $\uparrow$} & \multicolumn{1}{l}{KNN dist $\downarrow$} \\ \midrule
\multicolumn{5}{c}{\textbf{CelebA/CelebA/IR152}} \\ \hline 
KEDMI & - & - & 80.53 $\pm$ 3.86 & - & 1247.28 \\  
 + MA & 1 & EfficientNet-B0 & 81.20 $\pm$ 3.75 & 0.67 & 1234.16 \\ 
 + MA & 2 & EfficientNet-B0, EfficientNet-B1 & 84.47 $\pm$ 2.99 & 3.94 & 1223.56 \\  
 + MA & 3 & EfficientNet-B0, EfficientNet-B1, EfficientNet-B2 & 84.73 $\pm$ 3.76 & 4.20 & 1220.23 \\ 
 + MA & 4 & EfficientNet-B0, EfficientNet-B1, EfficientNet-B2, EfficientNet-B3 & \textbf{85.87 $\pm$ 2.63} & \textbf{5.34} & \textbf{1217.15} \\ \hline 
\end{tabular}
}

\label{tab:no_M_aug}
\end{table*}

\subsection{Different network architectures for \boldmath $M_{aug}$}
\label{sub:architecture_Maug}
In this section, we provide additional results by using different structures for augmenting the target model using $M_{aug}$ in the MI process. Note that the architecture of all these models is different from the one used for target model $M_t$.

More specifically, we use three different combinations for $M_{aug}$, each of which contains three models:
(i) \{EfficientNet-B0, EfficientNet-B1, EfficientNet-B2\}, and (ii) \{DenseNet121, DenseNet161, DenseNet169\}, and (iii) \{EfficientNet-B0, DenseNet121, MobileNetV3\}.
Results in
Table \ref{tab:architecture_M_aug} shows that $+ MA$ (Ours) consistently improves the attack accuracy and KNN distance with different network architectures.

\begin{table*}[t]
\caption{
We report top-1 attack accuracies, the improvement compared to the SOTA MI (Imp.), and KNN distance for different structures of network $M_{aug}$. Following exact experiment setups in \cite{chen2021knowledge}, here method = KEDMI, $\mathcal{D}_{priv}$ = CelebA, $\mathcal{D}_{pub}$ = CelebA, $M_t$ = IR152, evaluation model = face.evoLve. We select different network architectures for our experiment. Specifically, we use
Ours-1 = \{EfficientNet-B0\cite{tan2019efficientnet}, EfficientNet-B1\cite{tan2019efficientnet}, EfficientNet-B2\cite{tan2019efficientnet}\}, Ours-2 = \{DenseNet121 \cite{huang2017densely}, DenseNet161\cite{huang2017densely}, DenseNet169\cite{huang2017densely}\},  Ours-3 = \{EfficientNet-B0, DenseNet121\cite{huang2017densely}, MobileNetV3-large \cite{howard2019searching}\}. 
It shows that using different network architectures $M_{aug}$ consistently improves the attack accuracy and KNN distance.
}
\centering
\resizebox{0.9\width}{!}{%
\begin{tabular}{lcccc}
\toprule
Method & $M_{aug}$ & Attack Acc $\uparrow$ & Imp. $\uparrow$ & KNN dist $\downarrow$ \\ \midrule
\multicolumn{5}{c}{\textbf{CelebA/CelebA/IR152}} \\ \hline 
KEDMI & - & 80.53 $\pm$ 3.86 & - & 1247.28 \\
+ MA (Ours-1) & EfficientNet-B0, EfficientNet-B1, EfficientNet-B2 & \textbf{84.73 $\pm$ 3.76} & \textbf{4.20} & \textbf{1220.23}\\
+ MA (Ours-2) & DenseNet121, DenseNet161, DenseNet169 & \textbf{89.07 $\pm$ 3.32} & \textbf{8.54} & \textbf{1211.73}\\ 
+ MA (Ours-3) & EfficientNet-B0, DenseNet121, MobileNetV3-large & \textbf{86.53 $\pm$ 1.98} & \textbf{6.00} & \textbf{1204.94}\\ \hline 

\end{tabular}
}
\label{tab:architecture_M_aug}
\end{table*}

\subsection{The effect of different sizes of public dataset}
\label{sup:public_dataset}

We conduct additional experiments using different sizes of $\mathcal{D}_{pub}$ (10\%, 50\%) to emulate the different quality of prior information. 
The results for KEDMI \cite{chen2021knowledge} are shown in Table \ref{tab:celeba_dataset_size}.
The key observations are:
\begin{itemize}
    \item Baseline attack accuracies are poorer under limited $\mathcal{D}_{pub}$, \ie $\mathcal{D}_{pub}$ = 10\%.
    \item Our proposed method can outperform existing SOTA under varying degrees of prior information although the improvement obtained by KD is marginal under $\mathcal{D}_{pub}$ = 10\%.
\end{itemize}

\begin{table*}
\caption{
Sensitivity of the proposed method to prior information, $\mathcal{D}_{pub}$:
We use $\mathcal{D}_{priv}/ \mathcal{D}_{pub}$ = CelebA, $M_t$ = face.evoLve, evaluation = face.evoLve and KEDMI \cite{chen2021knowledge}.
We report top 1 MI attack accuracy and KNN distance using 10\%, 50\% and 100\% of $D_{pub}$. As GAN is trained on $D_{pub}$, it affects the baseline KEDMI and our proposed method. The results show that + LOM and + MA consistently improve upon the baseline. 
}
\centering

\begin{adjustbox}{width=0.99\width,center}{
\begin{tabular}{lcc|cc|cc}\toprule
&\multicolumn{2}{c}{\textbf{$\mathcal{D}_{pub}$ = 10\%}} &\multicolumn{2}{c}{\textbf{$\mathcal{D}_{pub}$ = 50\%}} &\multicolumn{2}{c}{\textbf{$\mathcal{D}_{pub}$ = 100\%}} \\
\midrule
&\textbf{Attack Acc $\uparrow$} &\textbf{KNN Dist $\downarrow$} &\textbf{Attack Acc $\uparrow$} &\textbf{KNN Dist $\downarrow$} &\textbf{Attack Acc $\uparrow$} &\textbf{KNN Dist $\downarrow$} \\
\midrule
KEDMI &58.33 $\pm$ 5.25 &1450.06 &79.07 $\pm$ 3.76 &1265.37 & 81.40 $\pm$ 3.25 & 1248.32 \\
+ LOM (Ours) &67.27 $\pm$ 1.83 &1395.38 &89.27 $\pm$ 0.96 &1202.45 &92.53 $\pm$ 1.51 & 1183.76 \\
+ MA (Ours) &61.80 $\pm$ 3.03 &1421.83 &82.20 $\pm$ 2.77 &1244.21 &85.07 $\pm$ 2.71 &1222.02 \\
+ LOMMA (Ours) &\textbf{74.40 $\pm$ 2.21} &\textbf{1328.79} &\textbf{89.67 $\pm$ 0.76} &\textbf{1170.37} &\textbf{93.20 $\pm$ 0.85} &\textbf{1154.32} \\
\bottomrule
\end{tabular}
}
\end{adjustbox}
\label{tab:celeba_dataset_size}
\end{table*}

\setcounter{figure}{0} 
\setcounter{table}{0} 

\section{Additional analysis and details on experimental setups}
\label{supp-sec:details}

\subsection{\boldmath Details on combining ${L}_{id}^{logit}$ and ${L}_{id}^{aug}$}
\label{sub:combination}

We provide details of combining ${L}_{id}^{logit}$ and ${L}_{id}^{aug}$. We substitute ${L}_{id}^{logit}$ (Eqn. 3 of main paper) into ${L}_{id}^{aug}$ (Eqn. 4 of main paper) for an inversion targeting class $k$ of the target model $M_t$, using augmented model $M_{aug}^{(i)}$. In particular, starting from Eqn. 4 of the main paper:
\begin{eqnarray}
{L}_{id}^{aug}(\bx;y) &=& 
   \gamma_t \cdot {L}_{id}(\bx;y,M_t)  \nonumber   \\
   && + \gamma_{aug} \cdot \sum_{i=1}^{N_{aug}} {L}_{id}(\bx;y,M_{aug}^{(i)})  \nonumber \\
 &=& 
   \gamma_t \cdot {L}_{id}^{logit}(\bx;y,M_t)  \nonumber   \\
   && + \gamma_{aug} \cdot \sum_{i=1}^{N_{aug}} {L}_{id}^{logit}(\bx;y,M_{aug}^{(i)})  \nonumber \\ 
 &=& 
   \gamma_t \cdot (-\pvector_t^T\wvector_{t,k} + \lambda ||\pvector_t - \pvector_{reg}||^2_2)  \nonumber   \\
   && + \gamma_{aug} \cdot \sum_{i=1}^{N_{aug}} (-   {(\pvector_{aug}^{(i)})^T(\wvector_{aug,k}^{(i)}) } \nonumber \\
   && + \lambda ||\pvector_{aug}^{(i)} - \pvector_{reg}||^2_2) \nonumber  \\
 &\approx& 
   \gamma_t \cdot (-\pvector_t^T\wvector_{t,k}  )  \nonumber   \\
   && + \gamma_{aug} \cdot \sum_{i=1}^{N_{aug}} (-(\pvector_{aug}^{(i)})^T(\wvector_{aug,k}^{(i)})) \nonumber \\
   && + \lambda' ||\pvector_{t} - \pvector_{reg}||^2_2 
   \label{eq:augmentation1}
\end{eqnarray}
Here, 
$\pvector_{t}$, 
$\wvector_{t,k}$
are penultimate layer activation and last layer weight for the target model $M_{t}$; 
$\pvector_{aug}^{(i)}$, 
$\wvector_{aug,k}^{(i)}$
are penultimate layer activation and last layer weight for the augmented model $M_{aug}^{(i)}$.
Note that one regularization is sufficient as shown in the last step. Eqn. \ref{eq:augmentation1} above is used in Eqn. 1 of the main paper in the inversion step using the proposed method.

\subsection{Details on improving KEDMI baseline}
\label{sub:zclipping}

\begin{table*}[t]
\caption{
We apply a simple technique that is introduced by GMI\cite{zhang2020secret} to get better baseline results for KEDMI \cite{chen2021knowledge}.
We report the results for KEDMI with and without $\bz$ clipping for IR152, face.evoLve, and VGG16 target model. Following exact experiment setups in \cite{chen2021knowledge}, here $\mathcal{D}_{priv}$ = CelebA, $\mathcal{D}_{pub}$ = CelebA, evaluation model = face.evoLve.
We report top-1 attack accuracies, the improvement compared to the SOTA MI (Imp.), and KNN distance. The improvement using $\bz$ clipping is clear. 
}

\centering
\resizebox{0.9\width}{!}{%
\begin{tabular}{lccc}
\toprule
Method & \multicolumn{1}{l}{Attack Acc $\uparrow$} & \multicolumn{1}{l}{Imp. $\uparrow$} & \multicolumn{1}{l}{KNN dist $\downarrow$} \\ \midrule
\multicolumn{4}{c}{\textbf{CelebA/CelebA/IR152}} \\ \hline 
KEDMI w/o $\bz$ clipping &  78.53 $\pm$ 3.45 & - & 1270.87 \\ 
KEDMI with $\bz$ clipping & \textbf{80.53 $\pm$ 3.86} & \textbf{2.00} & \textbf{1247.28} \\ \hline
\multicolumn{4}{c}{\textbf{CelebA/CelebA/face.evoLve}} \\ \hline
KEDMI w/o $\bz$ clipping  &  78.00 $\pm$ 4.09 & - & 1290.62 \\
KEDMI with $\bz$ clipping & \textbf{81.40 $\pm$ 3.25} &  \textbf{3.40} & \textbf{1248.32}\\\hline
\multicolumn{4}{c}{\textbf{CelebA/CelebA/VGG16}} \\ \hline
KEDMI w/o $\bz$ clipping  & 67.93 $\pm$ 4.24 & - & 1345.03 \\
KEDMI with $\bz$ clipping & \textbf{74.00 $\pm$ 3.10} & \textbf{6.07} & \textbf{1289.88}\\ \hline
\end{tabular}
}
\label{tab:clipz}
\end{table*}

We apply a simple technique that is introduced by GMI\cite{zhang2020secret} to get better results for KEDMI \cite{chen2021knowledge}. Specifically, after model inversion, and sampling $\bz$ from the learned distribution, we clip all elements of  $\bz$ into $[-1, 1]$, which is shown to be beneficial in \cite{zhang2020secret}.
In Table \ref{tab:clipz}, we observe that clipping $\bz$ help to boost the attack accuracy of KEDMI  and the reconstructed images are more similar to the private dataset as KNN distances are reduced. Therefore, for all the experiments with KEDMI in the main paper and Supp, we clip $\bz$ to get better results and we compare with this better version of KEDMI.

\def\pvector{\textbf{p}}
\def\wvector{\textbf{w}}

\subsection{Additional details on computing \boldmath $\pvector_{reg}$} 
\label{sub:pvector}
In Sec. 3.1, we propose an improved formulation for identity loss ${L}_{id}^{logit}$ which includes a regularization term $||\pvector - \pvector_{reg}||^2_2$ to prevent unbound growth of norm during optimization. Here we provide additional details on computing $\pvector_{reg}$.

Given that the attacker has no access to private training data, we estimate $\pvector_{reg}$ by a simple method using {\em public} data.
We firstly construct the set of penultimate layer features of public data using the target model and estimate the mean $\mu_{pen}$ and variance $\sigma_{pen}^2$:

\begin{equation}
    \mu_{pen} = \frac{1}{N} \sum^N_{i=1}M^{pen}(\bx_i)
\end{equation}
\begin{equation}
    \sigma_{pen}^2 =  \frac{1}{N} \sum^N_{i=1}(M^{pen}(\bx_i)-\mu_{pen})^2
\end{equation}
where $\bx_i$ is a sample from public dataset $\mathcal{D}_{pub}$, and $M^{pen}()$ operator returns the penultimate layer representations of the target model $M_t$ for a given input $\bx$. 
We analyze two ways to estimate $\pvector_{reg}$ as follow:
\begin{itemize}
    \item Fixed $\pvector_{reg}$ where  $\pvector_{reg}=\mu_{pen}$.
    \item $\pvector_{reg}$ is sampled using the prior distribution $\mathcal{N}(\mu_{pen}, \sigma_{pen})$.
\end{itemize}
Empirically, we use $N=5,000$ images from the public dataset $\mathcal{D}_{pub}$ to estimate $\mu_{pen}$ and $\sigma_{pen}$. The results show that 
using  $\pvector_{reg}$ which is sampled from $\mathcal{N}(\mu_{pen}, \sigma_{pen})$
gives better performance than using fixed $\pvector_{reg}={\mu_{pen}}$ (see 
Table \ref{tab:p_reg}). Therefore, all the results reported in the main paper use the $\pvector_{reg} \sim \mathcal{N}({\mu_{pen}}, \sigma_{pen})$.
We remark again that $\pvector_{reg}$ is estimated from {\em public} dataset.

\begin{table}[t]
\caption{
We report the results for KEDMI using 
a fixed $\pvector_{reg}$ or sampling from a distribution approximated for $\pvector_{reg}$. We use three different target models: IR152, face.evoLve, and VGG16. Following exact experiment setups in \cite{chen2021knowledge}, here $\mathcal{D}_{priv}$ = CelebA, $\mathcal{D}_{pub}$ = CelebA, evaluation model = face.evoLve.
We report top-1 attack accuracies, the improvement compared to the SOTA MI (Imp.), and KNN distance.}

\centering
\resizebox{0.9\columnwidth}{!}{%
\begin{tabular}{lcc}
\toprule
Method & \multicolumn{1}{l}{Attack Acc $\uparrow$} & \multicolumn{1}{l}{KNN dist $\downarrow$} \\ \midrule
\multicolumn{3}{c}{\textbf{CelebA/CelebA/IR152}} \\ \hline 
+ LOM (Fixed $p_{reg}$) & 92.27 $\pm$ 1.37 & \textbf{1155.92}\\ 
+ LOM (Ours) & \textbf{92.47 $\pm$ 1.41}  & 1168.55\\ \hline
\multicolumn{3}{c}{\textbf{CelebA/CelebA/face.evoLve}} \\ \hline
+ LOM (Fixed $p_{reg}$)& 90.40 $\pm$ 1.68 & 1257.95 \\ 
+ LOM (Ours) & \textbf{92.53 $\pm$ 1.51} &  \textbf{1183.76}\\\hline
\multicolumn{3}{c}{\textbf{CelebA/CelebA/VGG16}} \\ \hline
+ LOM (Fixed $p_{reg}$)& 85.60 $\pm$ 1.79 & 1259.60\\ 
+ LOM (Ours) & \textbf{89.07 $\pm$ 1.46} & \textbf{1218.46}\\\hline
\end{tabular}
}
\label{tab:p_reg}
\end{table}

\subsection{Details on regularization parameter \boldmath $\lambda$}
\label{sub:lambda}

In Sec 3.1 of the main paper, the regularization term $||\pvector - \pvector_{reg}||^2_2$ includes a parameter $\lambda$ which controls the effect of this term.
In this section, we evaluate the effect of this parameter  by examining different values of $\lambda$ on model inversion performance.
Results in Table \ref{tab:lambda} show that attack accuracy is improved over SOTA KEDMI 
with our proposed logit loss even
without the regularization term ($\lambda=0$). However, we get better results if the regularization is added \eg $\lambda = 1.0$. Due to its better performance, we use $\lambda = 1.0$ in all experiments with the proposed method.

\begin{table}[t]
\caption{We report the results for KEDMI with different $\lambda$ values using IR152 as target model. Following exact experimental setups in \cite{chen2021knowledge}, here $\mathcal{D}_{priv}$ = CelebA, $\mathcal{D}_{pub}$ = CelebA, evaluation model = face.evoLve.
We report top-1 attack accuracies, the improvement compared to the SOTA MI (Imp.), and KNN distance.
}
\centering
\resizebox{0.95\columnwidth}{!}{%
\begin{tabular}{lcccc}
\toprule
Method & $\lambda$ & Attack Acc $\uparrow$ & Imp. $\uparrow$ & KNN dist $\downarrow$ \\ \midrule
\multicolumn{5}{c}{\textbf{CelebA/CelebA/IR152}} \\ \hline 
KEDMI & - & 80.53 $\pm$ 3.86 & - & 1247.28 \\
 + LOM & 0 & 90.33 $\pm$ 1.64 & 9.80 & 1198.39 \\
 + LOM & 0.5 & 89.53 $\pm$ 1.21 & 9.00 & 1175.35 \\
 + LOM & 1.0 & \textbf{92.47 $\pm$ 1.41} & \textbf{11.94} & 1168.55 \\
 + LOM & 2.0 & 91.87 $\pm$ 1.09 & 11.34 & 1125.54 \\
 + LOM & 10.0 & 85.80 $\pm$ 1.24 & 5.27 & \textbf{1110.80} \\ \hline

\end{tabular}
}
\label{tab:lambda}
\end{table}

\subsection{Computational overhead }

In order to investigate the computational overhead introduced by our proposed method, in this section,
we report the running time for reconstructing images of 300 identities on CelebA/CelebA/IR152 setup for KEDMI and GMI, and 100 identities on CelebA/CelebA/ResNet-34 for VMI.  
All the experiments of KEDMI and GMI are performed on an NVIDIA GeForce RTX 3090 GPU, and the experiments of VMI are performed on an NVIDIA RTX A5000 GPU.
The results in Table \ref{tab:runtime} show that + LOM does not affect the training time compared to the baseline. However, + MA adds some computational overhead as it uses additional networks $M_{aug}$ during the inversion. 

\begin{table}[h]
\caption{Computational complexity of different algorithms in terms of average running time (GPU hours) using single GPU. We use KEDMI, GMI, and VMI approaches as the baseline. We have also included the running time Ratio when compared to the corresponding baseline.}
\centering
\resizebox{0.7\columnwidth}{!}{%
\begin{tabular}{lll}
\toprule
{\bf Method} & {\bf RunTime (hrs)} & {\bf Ratio} \\

\midrule
KEDMI & 0.35 & 1.00 \\
+ LOM (Ours) & 0.35 & 1.00
\\
+ MA (Ours) & 0.60 & 1.71\\
+ LOMMA (Ours) & 0.60 & 1.72\\
\midrule

GMI & 1.61 & 1.00\\
+ LOM (Ours) & 1.61 & 1.00
 \\
+ MA (Ours) & 2.83 & 1.76
 \\
+ LOMMA (Ours) & 2.85 & 1.77
 \\
\midrule

VMI & 364.67 & 1.00\\
+ LOM (Ours) & 368.24 & 1.01\\
+ MA (Ours) & 368.69 & 1.01 \\
+ LOMMA (Ours) & 379.41 & 1.04\\
\bottomrule
\end{tabular}
}
\label{tab:runtime}
\end{table}

\subsection{Hyperparameters}
In the experiments of GMI and KEDMI, we do the inversion using SGD optimizer with the learning rate $lr = 0.02$ in 2400 iterations which are used from the released code of KEDMI \footnote{https://github.com/SCccc21/Knowledge-Enriched-DMI}.
We set $\gamma_t = \gamma_{aug} = 100/(N_{aug}+1)$ and $\lambda=100$, where $N_{aug}$ is the number of models used for augmented model $M_{aug}$. 
We estimate $\pvector_{reg}$  for each classifier by using $N=5,000$ images from the public dataset $\mathcal{D}_{pub}$.
In the experiments of VMI, we use 20 epochs (equal to 3120 iterations) to learn the distribution of each identity.

\subsection{Dataset}

\textbf{Experiments of KEDMI and GMI.}
We follow exact experimental setups in \cite{chen2021knowledge}.
For the CelebA task, we use the dataset divided by \cite{chen2021knowledge} for all of the experiments. In particular, the private dataset has 30,027 images of 1000 identities and the public dataset has 30000 images that are non-overlapping identities with the private dataset. In the experiments in Table 5 (main paper), we use FFHQ \cite{karras2019style} as the public dataset to train GAN and distill knowledge to augmented models. For MNIST and CIFAR-10 tasks, the private dataset contains images with labels from 0 to 4 and the public dataset includes the rest of the dataset with labels from 5 to 9.

\textbf{Experiments of VMI.}
We follow exact experimental setup in \cite{wang2021variational}.
We use the CelebA dataset and MNIST dataset for VMI experiments. For CelebA, we follow \cite{wang2021variational} to divide the dataset into two parts. The first part contains images of 1000 most frequent identities which uses as private dataset. The rest of dataset is used as public dataset. For the experiments on MNIST dataset, we use EMNIST\cite{cohen2017emnist} as public dataset to train GAN and $M_{aug}$.

\setcounter{figure}{0} 
\setcounter{table}{0}

\section{Additional Visualizations}
\label{supp-sec:visualization}

\subsection{Additional Results for GMI}
\label{supp-subsec:gmi-qualitative-results}
\begin{figure} [ht]
\begin{adjustbox}{width=0.99\columnwidth,left}
\begin{tabular}{c}
    \includegraphics[width=0.99\columnwidth]{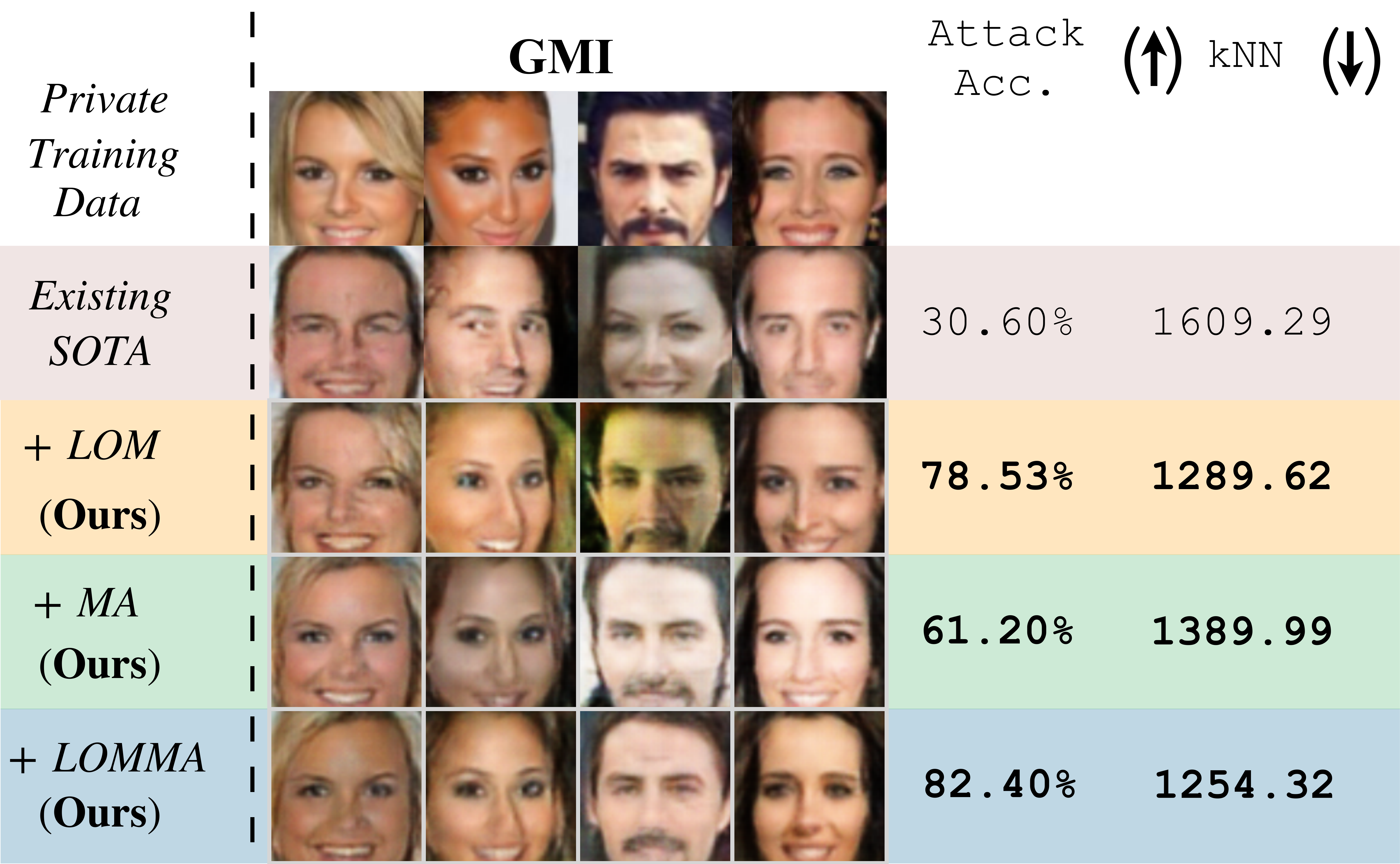}
\end{tabular}
\end{adjustbox}
\caption{
Qualitative / Quantitative (Top1 Attack Acc., KNN Dist) results to demonstrate the efficacy of our proposed method.
We use GMI \cite{zhang2020secret}, $\mathcal{D}_{priv}$ = CelebA \cite{liu2015deep}, $\mathcal{D}_{pub}$ = CelebA \cite{liu2015deep} and $M$ = IR152 \cite{he2016deep}.
As one can observe, our proposed method achieves better reconstruction of private data both visually and quantitatively (validated by KNN results) resulting in a significant boost in attack performance.
}
\label{supp-fig:gmi_qualitative_study}
\end{figure}

Similar to results reported for KEDMI (Figure \textcolor{red}{4}, main paper), in this section, we show results for GMI \cite{zhang2020secret} under IR152 target classifier to show the efficacy of our proposed methods. The result is shown in Figure \ref{supp-fig:gmi_qualitative_study}.

\subsection{Penultimate layer visualization for GMI, KEDMI and VMI}
\label{supp-subsec:penultimate_visualization}

\begin{figure*}[ht]
\begin{adjustbox}{width=0.99\textwidth,center}
\begin{tabular}{c}
    \includegraphics[width=0.99\textwidth]{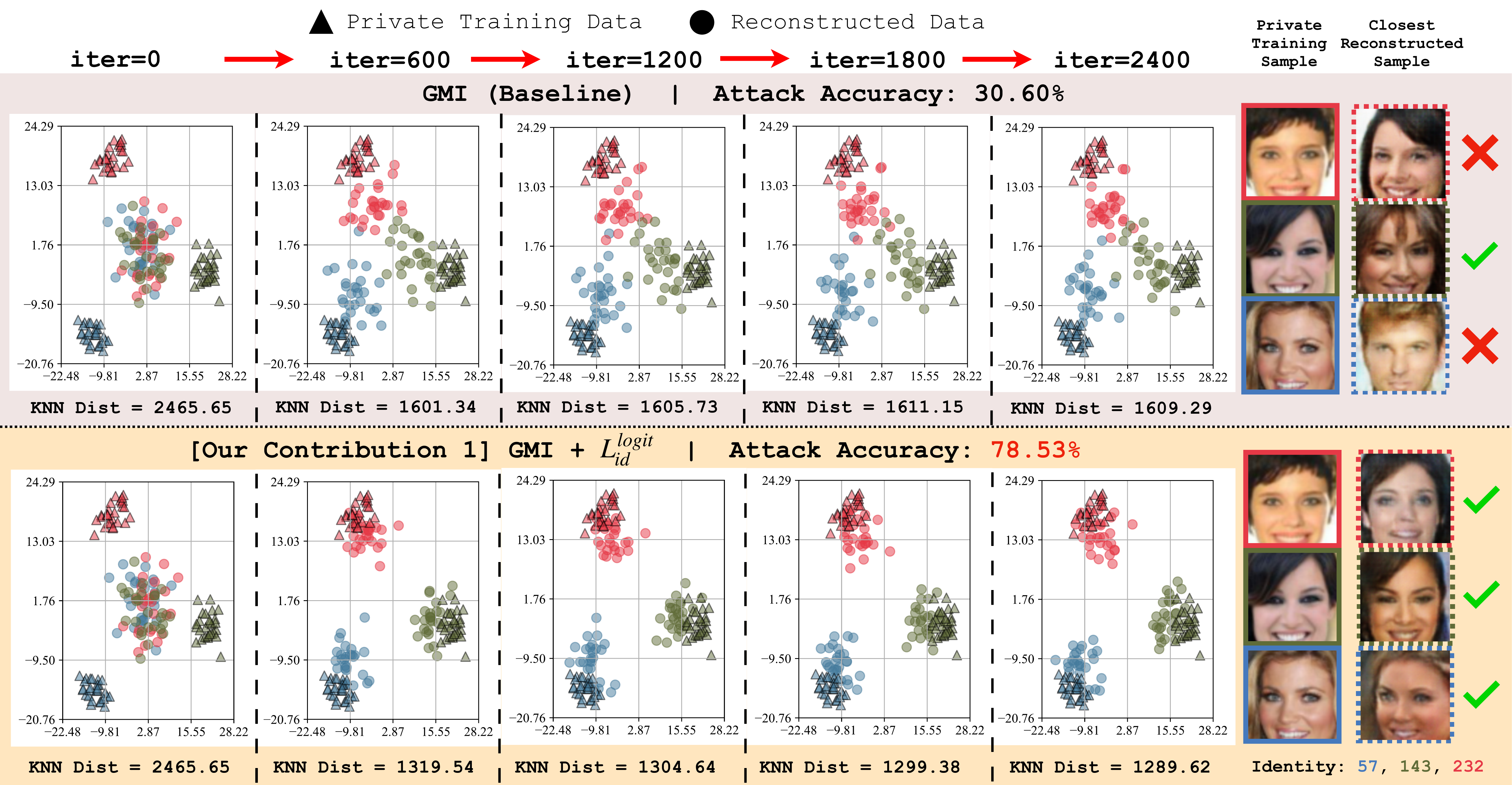}
\end{tabular}
\end{adjustbox}
\caption
{
Visualization of the penultimate layer representations ($\mathcal{D}_{priv}$ = CelebA \cite{liu2015deep}, $\mathcal{D}_{pub}$ = CelebA \cite{liu2015deep}, $M_t$ = IR152 \cite{he2016deep}, Evaluation Model = face.evoLve \cite{yu_know_you_at_one_glance}, Inversion iterations = 2400) for private training data and reconstructed data using GMI \cite{zhang2020secret}.
Following exact evaluation protocol in \cite{chen2021knowledge}, we use face.evoLve \cite{yu_know_you_at_one_glance} to extract representations.
We show results for 3 randomly chosen identity.
We include KNN distance (for different iterations) and final attack accuracy following the protocol in \cite{chen2021knowledge}.
For each identity, we also include a randomly selected private training data and the closest reconstructed sample at iteration=2400.
\textbf{\textcircled{\raisebox{-0.8pt}{1}}
Identity loss in SOTA MI methods \cite{zhang2020secret, chen2021knowledge, wang2021variational}
(Eqn. \textcolor{red}{2}, main paper)
is sub-optimal for MI (Top).}
Using penultimate representations during inversion, we observe 2 instances (\eg target identity {\bf \color[HTML]{4579BD} 57} and {\bf \color[HTML]{E63946} 232}) 
where GMI \cite{zhang2020secret} (using Eqn. \textcolor{red}{2}, main paper for identity loss) is unable to reconstruct data close to private training data.
Hence, private and reconstructed facial images are qualitatively different. 
\textbf{\textcircled{\raisebox{-0.8pt}{2}} 
Our proposed identity loss, ${L}_{id}^{logit}$ (Eqn. \textcolor{red}{3}, main paper), 
can 
effectively guide reconstruction of data close to 
private training data 
(Bottom)}.
This can be clearly observed using both penultimate layer representations and KNN distances for all 3 target classes {\bf \color[HTML]{4579BD} 57}, {\bf \color[HTML]{606c38} 143} and {\bf \color[HTML]{E63946} 232}.
Best viewed in color.
}
\label{fig-supp:gmi_penultimate_visualization}
\end{figure*}

\begin{figure*}[ht]
\begin{adjustbox}{width=0.99\textwidth,center}
\begin{tabular}{c}
    \includegraphics[width=0.99\textwidth]{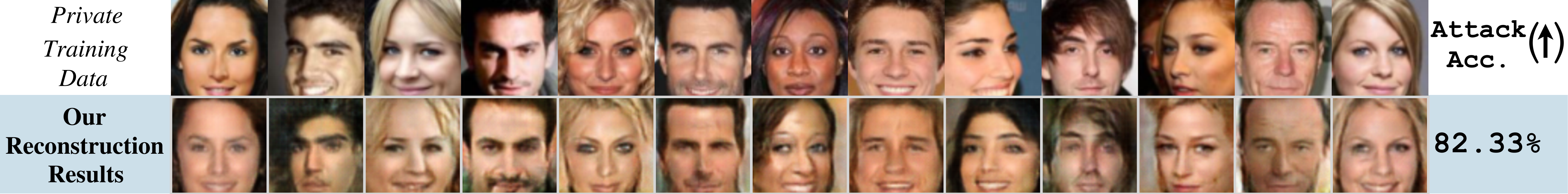}
\end{tabular}
\end{adjustbox}
\caption
{
We show private data (top), \textit{our} reconstruction results (bottom) and Attack accuracy
($\mathcal{D}_{priv}$ = CelebA \cite{liu2015deep}, $\mathcal{D}_{pub}$ = CelebA \cite{liu2015deep}, $M_t$ = face.evoLve \cite{yu_know_you_at_one_glance}, Evaluation Model = face.evoLve \cite{yu_know_you_at_one_glance}, Inversion iterations = 2400) using GMI \cite{zhang2020secret}.
We remark that these results are obtained by combining ${L}_{id}^{logit}$ and ${L}_{id}^{aug}$ (referred to as \textbf{+ LOMMA} throughout the paper).
}
\label{fig-supp:celeba_celeba_face_evolve_gmi}
\end{figure*}
\begin{figure*}[ht]
\begin{adjustbox}{width=0.99\textwidth,center}
\begin{tabular}{c}
    \includegraphics[width=0.99\textwidth]{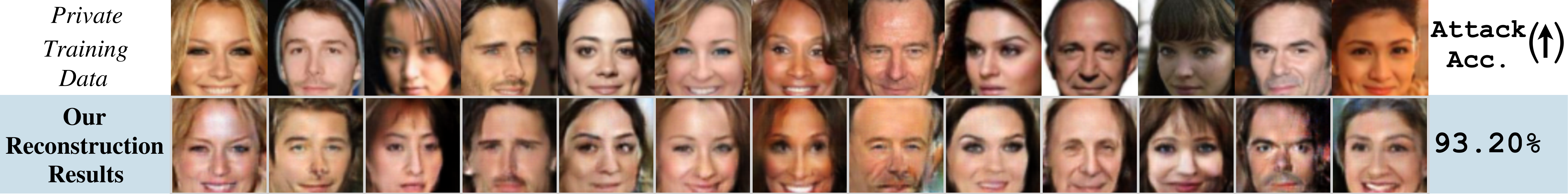}
\end{tabular}
\end{adjustbox}
\caption
{
We show private data (top), \textit{our} reconstruction results (bottom) and Attack accuracy
($\mathcal{D}_{priv}$ = CelebA \cite{liu2015deep}, $\mathcal{D}_{pub}$ = CelebA \cite{liu2015deep}, $M_t$ = face.evoLve \cite{yu_know_you_at_one_glance}, Evaluation Model = face.evoLve \cite{yu_know_you_at_one_glance}, Inversion iterations = 2400) using KEDMI \cite{chen2021knowledge}.
We remark that these results are obtained by combining ${L}_{id}^{logit}$ and ${L}_{id}^{aug}$ (referred to as \textbf{+ LOMMA} throughout the paper).
We remark that in the standard CelebA benchmark, our method boosts attack accuracy significantly, achieving more than 90\% attack accuracy for the first time in contemporary MI literature.
}
\label{fig-supp:celeba_celeba_face_evolve_kedmi}
\vspace{-0.5cm}
\end{figure*}

\begin{figure*}[!t]
\begin{adjustbox}{width=0.99\textwidth,center}
\begin{tabular}{c}
    \includegraphics[width=0.99\textwidth]{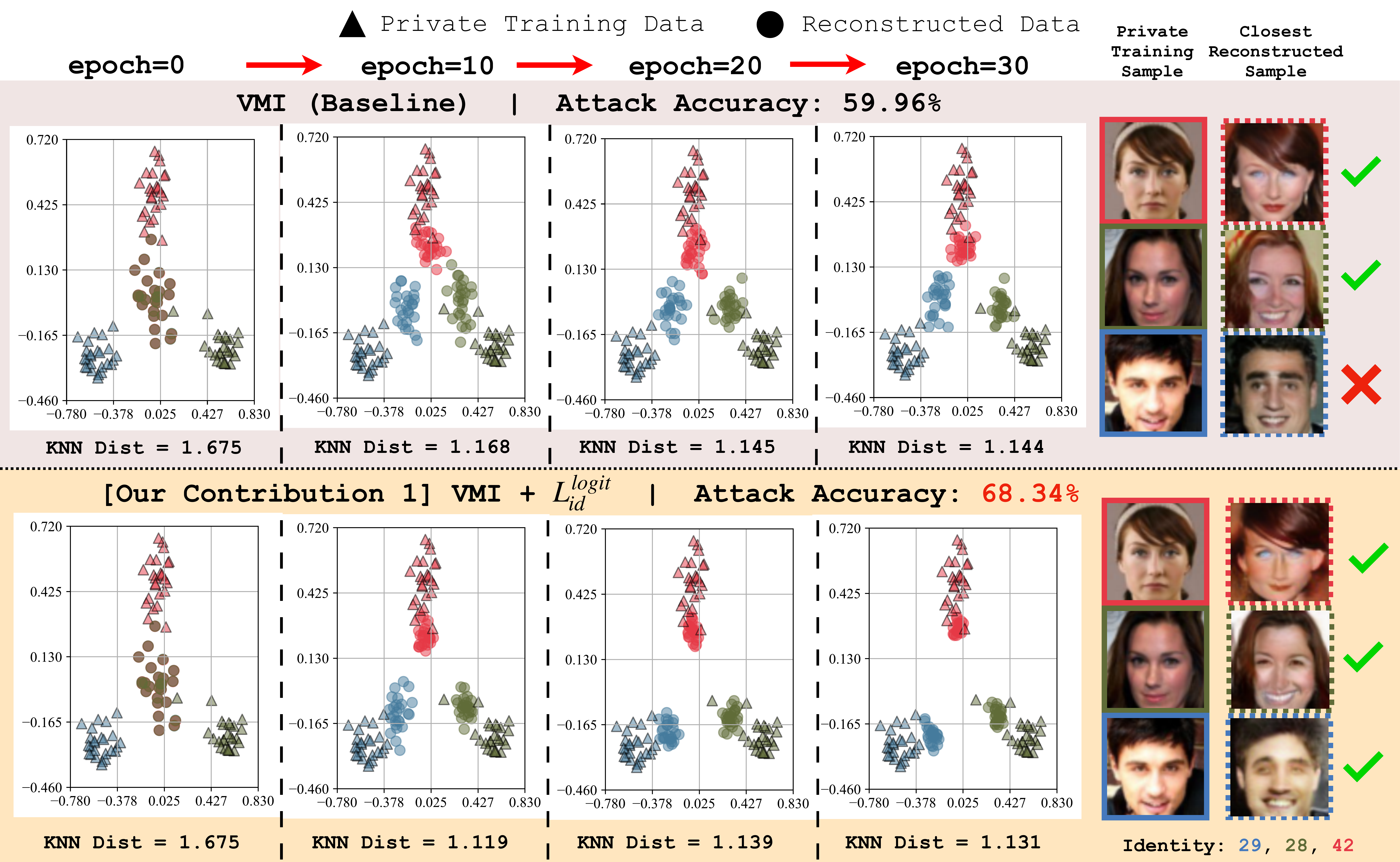}
\end{tabular}
\end{adjustbox}
\caption
{
Visualization of the penultimate layer representations ($\mathcal{D}_{priv}$ = CelebA \cite{liu2015deep}, $\mathcal{D}_{pub}$ = CelebA \cite{liu2015deep}, $M_t$ = ResNet34 \cite{wang2021variational}, Evaluation Model = IR-SE50 \cite{wang2021variational}, Inversion epochs = 30) for private training data and reconstructed data using VMI \cite{wang2021variational}.
Following exact evaluation protocol in \cite{wang2021variational}, we use IR-SE50 to extract representations.
We show results for 3 randomly chosen identity.
We include KNN distance 
and final attack accuracy. 
Given that we strictly follow \cite{wang2021variational}, we remark that due to the use of IR-SE50 evaluation classifier to extract penultimate layer representations, the features have different scales resulting in lower KNN distances (compared to KEDMI and GMI results).
For each identity, we include a randomly selected private training data and the closest reconstructed sample (epoch = 30).
\textbf{\textcircled{\raisebox{-0.8pt}{1}}
Identity loss in SOTA MI methods \cite{zhang2020secret, chen2021knowledge, wang2021variational}
(Eqn. \textcolor{red}{2}, main paper)
is sub-optimal for MI (Top).}
Using penultimate representations during inversion, we observe an instance
(\eg target identity {\bf \color[HTML]{4579BD}29}) 
where VMI \cite{wang2021variational} (using Eqn. \textcolor{red}{2}, main paper for identity loss) is unable to reconstruct data close to private training data.
Hence, private and reconstructed facial images are qualitatively different. 
\textbf{\textcircled{\raisebox{-0.8pt}{2}} 
Our proposed identity loss, ${L}_{id}^{logit}$ (Eqn. \textcolor{red}{3}, main paper), 
can 
effectively guide reconstruction of data close to 
private training data 
(Bottom)}.
This can be observed using penultimate layer representations and KNN distances for all 3 target classes {\bf \color[HTML]{4579BD} 29}, {\bf \color[HTML]{606c38} 28} and {\bf \color[HTML]{E63946} 42}.
Best viewed in color.
}
\label{fig-supp:vmi_penultimate_visualization}
\end{figure*}

\begin{figure*}[!t]
\begin{adjustbox}{width=0.99\textwidth,center}
\begin{tabular}{c}
    \includegraphics[width=0.99\textwidth]{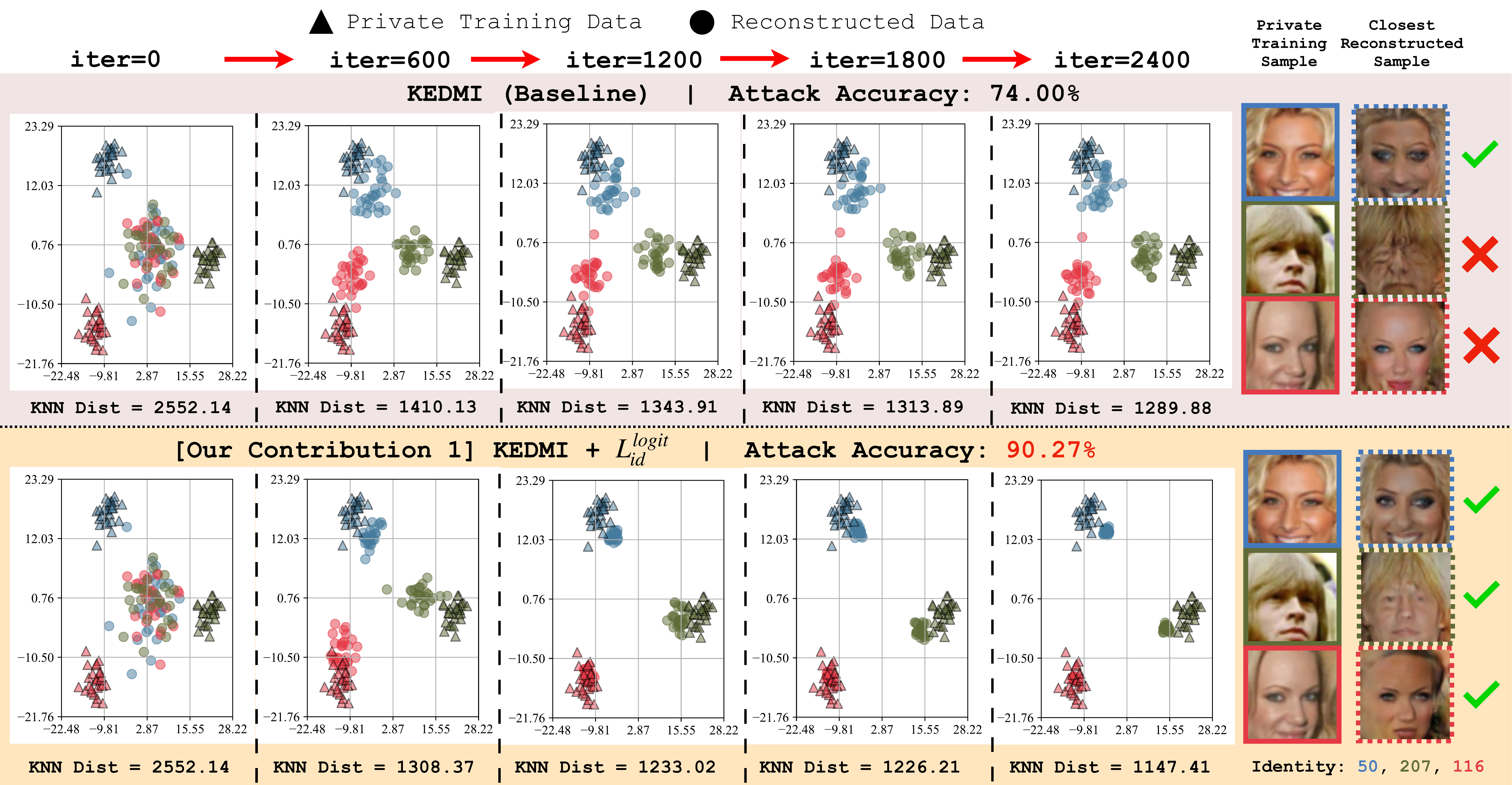}
\end{tabular}
\end{adjustbox}
\caption
{
Visualization of the penultimate layer representations ($\mathcal{D}_{priv}$ = CelebA \cite{liu2015deep}, $\mathcal{D}_{pub}$ = CelebA \cite{liu2015deep}, $M_t$ = VGG16 \cite{simonyan2014very}, Evaluation Model = face.evoLve \cite{yu_know_you_at_one_glance}, Inversion iterations = 2400) for private training data and reconstructed data using KEDMI \cite{chen2021knowledge}.
Following exact evaluation protocol in \cite{chen2021knowledge}, we use face.evoLve \cite{yu_know_you_at_one_glance} to extract representations.
We show results for 3 randomly chosen identity.
We include KNN distance (for different iterations) and final attack accuracy following the protocol in \cite{chen2021knowledge}.
For each identity, we also include a randomly selected private training data and the closest reconstructed sample at iteration=2400.
\textbf{\textcircled{\raisebox{-0.8pt}{1}}
Identity loss in SOTA MI methods \cite{zhang2020secret, chen2021knowledge, wang2021variational}
(Eqn. \textcolor{red}{2}, main paper)
is sub-optimal for MI (Top).}
Using penultimate representations during inversion, we observe 2 instances (\eg target identity {\bf \color[HTML]{606c38} 207} and {\bf \color[HTML]{E63946} 116}) 
where KEDMI \cite{chen2021knowledge} (using Eqn. \textcolor{red}{2}, main paper for identity loss) is unable to reconstruct data close to private training data.
Hence, private and reconstructed facial images are qualitatively different. 
\textbf{\textcircled{\raisebox{-0.8pt}{2}} 
Our proposed identity loss, ${L}_{id}^{logit}$ (Eqn. \textcolor{red}{3}, main paper), 
can 
effectively guide reconstruction of data close to 
private training data 
(Bottom)}.
This can be clearly observed using both penultimate layer representations and KNN distances for all 3 target classes {\bf \color[HTML]{4579BD} 50}, {\bf \color[HTML]{606c38} 207} and {\bf \color[HTML]{E63946} 116}.
Best viewed in color.
}
\label{fig:kedmi_penultimate_visualization_face_evolve_target}
\end{figure*}

In this section, we show additional penultimate layer visualizations to support our formulation of ${L}_{id}^{logit}$ as an improved MI Identity Loss.
We show visualizations for GMI \cite{zhang2020secret} and VMI \cite{wang2021variational} in Figures \ref{fig-supp:gmi_penultimate_visualization} and \ref{fig-supp:vmi_penultimate_visualization} respectively.
Further, we show penultimate layer visualization for an additional target classifier, face.evoLve using KEDMI \cite{chen2021knowledge} in Figure \ref{fig:kedmi_penultimate_visualization_face_evolve_target} to validate our findings.

\begin{figure*}
\begin{adjustbox}{width=0.99\textwidth,center}
\begin{tabular}{c}
    \includegraphics[width=0.99\textwidth]{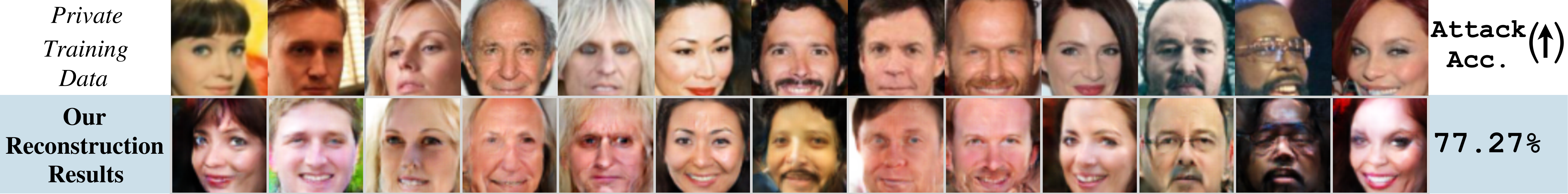}
\end{tabular}
\end{adjustbox}
\caption
{
\textit{Cross-dataset MI results}.
We show private data (top), \textit{our} reconstruction results (bottom) and Attack accuracy
($\mathcal{D}_{priv}$ = CelebA \cite{liu2015deep}, $\mathcal{D}_{pub}$ = FFHQ \cite{karras2019style}, $M_t$ = IR152 \cite{he2016deep}, Evaluation Model = face.evoLve \cite{yu_know_you_at_one_glance}, Inversion iterations = 2400) using KEDMI \cite{chen2021knowledge}.
Cross-dataset MI is a challenging setup due to the large distribution shift between private and public data.
We remark that these results are obtained by combining ${L}_{id}^{logit}$ and ${L}_{id}^{aug}$ (referred to as \textbf{+ LOMMA} throughout the paper).
}
\label{fig-supp:cross-dataset}
\end{figure*}

\begin{figure*}
\begin{adjustbox}{width=0.99\textwidth,center}
\begin{tabular}{c}
    \includegraphics[width=0.99\textwidth]{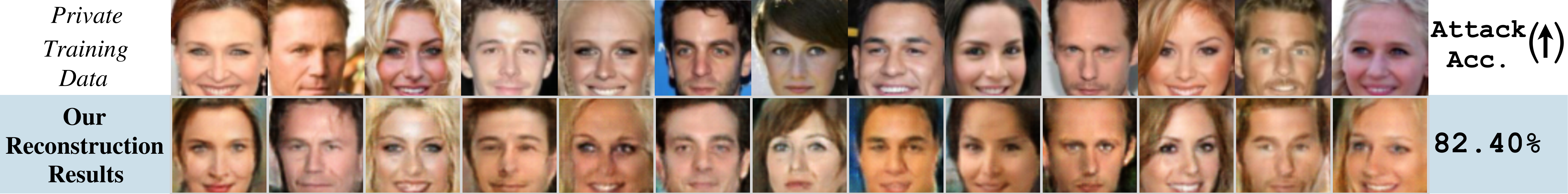}
\end{tabular}
\end{adjustbox}
\caption
{
We show private data (top), \textit{our} reconstruction results (bottom) and Attack accuracy
($\mathcal{D}_{priv}$ = CelebA \cite{liu2015deep}, $\mathcal{D}_{pub}$ = CelebA \cite{liu2015deep}, $M_t$ = IR152 \cite{he2016deep}, Evaluation Model = face.evoLve \cite{yu_know_you_at_one_glance}, Inversion iterations = 2400) using GMI \cite{zhang2020secret}.
We remark that these results are obtained by combining ${L}_{id}^{logit}$ and ${L}_{id}^{aug}$ (referred to as \textbf{+ LOMMA} throughout the paper).
}
\label{fig-supp:celeba_celeba_ir152_gmi}
\end{figure*}

\subsection{Our reconstruction results}
\label{supp-subsec:recon_results}

\begin{figure*}[!h]
\begin{adjustbox}{width=0.99\textwidth,center}
\begin{tabular}{c}
    \includegraphics[width=0.99\textwidth]{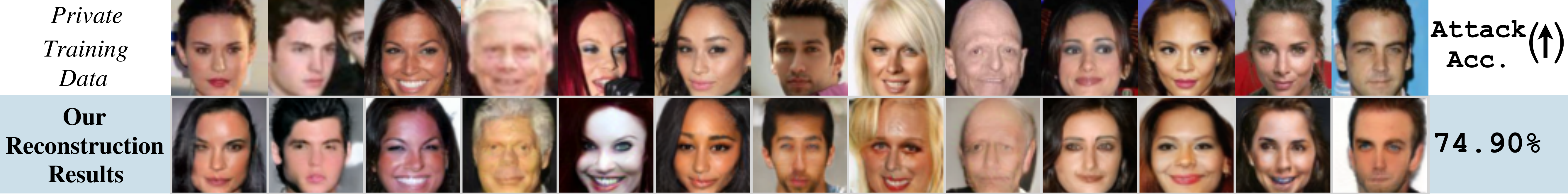}
\end{tabular}
\end{adjustbox}
\caption
{
We show private data (top), \textit{our} reconstruction results (bottom) and Attack accuracy
($\mathcal{D}_{priv}$ = CelebA \cite{liu2015deep}, $\mathcal{D}_{pub}$ = CelebA \cite{liu2015deep}, $M_t$ = ResNet34 \cite{wang2021variational}, Evaluation Model = IR-SE50 \cite{deng2019arcface}, Inversion epochs = 30) using VMI \cite{wang2021variational}.
We remark that these results are obtained by combining ${L}_{id}^{logit}$ and ${L}_{id}^{aug}$ (referred to as \textbf{+ LOMMA} throughout the paper).
}
\label{fig-supp:celeba_celeba_R34_vmi}
\end{figure*}

Given that the goal of MI is to reconstruct private training data, in this section, we show reconstructed samples for 5 additional setups using our proposed method.
We show reconstruction results using GMI \cite{zhang2020secret} and VMI \cite{wang2021variational} in Figures \ref{fig-supp:celeba_celeba_ir152_gmi} and \ref{fig-supp:celeba_celeba_R34_vmi} respectively.
Further, we show additional reconstruction results for GMI and KEDMI using a different target classifier (face.evoLve) in Figures \ref{fig-supp:celeba_celeba_face_evolve_gmi} and \ref{fig-supp:celeba_celeba_face_evolve_kedmi} to validate the efficacy of our proposed method.
Finally, we show reconstruction results for Cross-dataset MI in Figure \ref{fig-supp:cross-dataset}. We remark that cross-dataset MI is a challenging attack setup due to large distribution shift between private and public data.
Following \cite{wang2021variational}, we use FFHQ \cite{karras2019style} as the public dataset.
To conclude, we remark that the samples reconstructed using our proposed method closely resembles the private training data in many instances, and this is quantitatively validated using MI attack accuracy.

\setcounter{figure}{0} 
\setcounter{table}{0} 

 \section{Additional Related work}
 \label{supp-sec:related_work}
Given a trained model, Model Inversion (MI) aims to extract information about training data. 
Fredrikson et al. \cite{fredrikson2014privacy} propose one of the first methods for MI. The authors found that attackers can extract genomic and demographic information about patients using the ML model. 
In \cite{fredrikson2015model}, Fredrikson et al. extended the problem to the facial recognition setup where the authors can recover the face images.
In \cite{yang2019neural}, Yang et al. proposed adversarial model inversion which uses the target classifier as an encoder to produce a prediction vector. A second network takes the prediction vector as the input to reconstruct the data.

Instead of performing MI attacks directly on high-dimensional space (e.g. image space), recent works have proposed to reduce the search space to latent space by training a deep generator \cite{zhang2020secret,wang2021variational,chen2021knowledge,yang2019adversarial}.
In particular, a generator is trained on an auxiliary dataset that has a similar structure to the target image space.
In \cite{zhang2020secret}, the authors proposed GMI which uses a pretrained GAN to learn the image structure of the auxiliary dataset and finds the inversion images through the latent vector of the generator.
Chen et al. \cite{chen2021knowledge} extend GMI by training discriminator to distinguish the real and fake samples and to be able to predict the label as the target model. Furthermore, the authors proposed modeling the latent distribution to reduce the inversion time and improve the quality of reconstructed samples.
VMI \cite{wang2021variational} provides a probabilistic interpretation for MI and proposes a variational objective to approximate the latent space of target data.

Zhao et al.
\cite{zhao2021exploiting} propose to embed the information of model explanations for model inversion. A model explanation is trained to analyze and constrain the inversion model to learn useful activations. 
Another MI attack type is called label-only MI attacks which attackers only access the predicted label without a confidence probability \cite{choquette2021label,kahla2022label}.
Recently, Kahla et al. \cite{kahla2022label} propose to estimate the direction to reach the target class’s centroid for an MI attack.
In this work, we instead focus on a different problem and propose two improvements to the identity loss which is common among all SOTA MI approaches.
In future work, we hope to explore model inversion for different  tasks including multimodal learning and data-centric applications \cite{he2020momentum, Chandrasegaran_2021_CVPR, Chandrasegaran_2022_ECCV, tran2021data_aug_gan, lim2018doping, radford2021learning, koh2023grounding, zhang2021cross,  khosla2020supervised}.

\end{document}